\documentclass[lettersize,journal]{IEEEtran}
\usepackage{amsmath,amssymb,amsfonts}
\usepackage{array}
\usepackage[caption=false,font=scriptsize,labelfont=rm,textfont=rm]{subfig}
\usepackage{textcomp}
\usepackage{stfloats}
\usepackage{url}
\usepackage{verbatim}
\usepackage{graphicx}
\usepackage{cite}
\usepackage{bm}
\usepackage{xcolor}
\usepackage{pifont}
\usepackage{fancyhdr}
\usepackage{multirow}
\usepackage[ruled,linesnumbered]{algorithm2e}
\usepackage{booktabs}
\usepackage{epstopdf}

\def\BibTeX{{\rm B\kern-.05em{\sc i\kern-.025em b}\kern-.08em
    T\kern-.1667em\lower.7ex\hbox{E}\kern-.125emX}}
\begin{document}
\title{A Progressive Image Restoration Network for High-order Degradation Imaging in Remote Sensing}
\author{Yujie Feng, Yin Yang, Xiaohong Fan, Zhengpeng Zhang, Lijing Bu and Jianping Zhang

\thanks{This work was supported by the National Key Research and Development Program of China (2020YFA0713503), the science and technology innovation Program of Hunan Province (2024RC9008), the Project of Scientific Research Fund of the Hunan Provincial Science and Technology Department (2022RC3022, 2023GK2029, 2024ZL5017, 2024JJ1008), and Program for Science and Technology Innovative Research Team in Higher Educational Institutions of Hunan Province of China. Corresponding authors: J. Zhang (e-mail: jpzhang@xtu.edu.cn) and Y. Yang (e-mail: yangyinxtu@xtu.edu.cn).}
\thanks{Y. Feng is with the School of Mathematics and Computational Science, Xiangtan University, Hunan Key Laboratory for Computation and Simulation in Science and Engineering, Key Laboratory for Intelligent Computing and Information Processing of the Ministry of Education, Xiangtan 411105, China.}
\thanks{Y. Yang and J. Zhang are with the School of Mathematics and Computational Science, Xiangtan University, National Center for Applied Mathematics in Hunan, Hunan International Scientific and Technological Innovation Cooperation Base of Computational Science, Xiangtan 411105, China.}
\thanks{X. Fan is with the College of Mathematical Medicine, Zhejiang Normal University, Jinhua 321004, China.}
\thanks{Z. Zhang and L. Bu is with the School of Automation and Electronic Information, Xiangtan University, Xiangtan 411105, China.}
}

\markboth{Accepted to IEEE Transactions on Geoscience and Remote Sensing, July, 2025. \hfill DOI:10.1109/TGRS.2025.3590753}%
{How to Use the IEEEtran \LaTeX \ Templates}

\maketitle

\begin{abstract}
Recently, deep learning methods have gained remarkable achievements in the field of image restoration for remote sensing (RS). However, most existing RS image restoration methods focus mainly on conventional first-order degradation models, which may not effectively capture the imaging mechanisms of remote sensing images. Furthermore, many RS image restoration approaches that use deep learning are often criticized for their lacks of architecture transparency and model interpretability. To address these problems, we propose a novel progressive restoration network for high-order degradation imaging (HDI-PRNet), to progressively restore different image degradation. HDI-PRNet is developed based on the theoretical framework of degradation imaging, also Markov properties of the high-order degradation process and Maximum a posteriori (MAP) estimation, offering the benefit of mathematical interpretability within the unfolding network. The framework is composed of three main components: a module for image denoising that relies on proximal mapping prior learning, a module for image deblurring that integrates Neumann series expansion with dual-domain degradation learning, and a module for super-resolution. Extensive experiments demonstrate that our method achieves superior performance on both synthetic and real remote sensing images.
\end{abstract}

\section{Introduction}
\IEEEPARstart{R}{emote} sensing images contained abundant spatial texture details and scene semantic information, which have been widely applied in tasks such as scene segmentation \cite{wang2022unetformer}. However, due to the complex environment of remote sensing imaging and the inherent limitations of sensor hardware, the observed remote sensing images inevitably degrade. The low quality of remote sensing images limits their practical application in high-level remote sensing tasks directly. To address this problem efficiently with minimal cost, employing image restoration techniques is optimal, as it avoids the need for upgrading equipment. These techniques reconstruct a high-resolution (HR) image, enriched with detailed high-frequency textures, from a low-resolution (LR) image. As a result, image restoration has become increasingly popular in the fields of remote sensing and computer vision.

The classical first-order image degradation model can be expressed mathematically as follows
\begin{equation}
\begin{split}
\boldsymbol{g} &= (\boldsymbol{h}*\boldsymbol{u})_{\downarrow_{\boldsymbol{s}}} + \boldsymbol{n},
\end{split}
\label{eq66}
\end{equation}
where $\boldsymbol{g}$ is the observational LR image, $\boldsymbol{u}$ is the HR image with sharp details, $\boldsymbol{h}$ is the blur kernel, $\boldsymbol{n}$ is the additive white gaussian noise (AWGN), and $*$ denotes the 2-D convolution operator, $\downarrow_{\boldsymbol{s}}$ denotes the downsampling operator with scale factor $\boldsymbol{s}$.

The restoration of degradation images in remote sensing is a challenging task, 
as it involves restoring the sharp HR image $\boldsymbol{u}$ from the deteriorated LR image $\boldsymbol{g}$. This problem (\ref{eq66}) is considered ill-posed because there are multiple HR solutions that can map to the same LR image, resulting in a non-unique solution. Various methods \cite{zamir2022learning, cui2023focal, cho2021rethinking, cui2023selective, XHFan2023b, cui2024omni} for image restoration have been investigated. Although many algorithms ignore blur and noise degradation in the forward model \eqref{eq66} and rely solely on the ideal bicubic operator to simulate the super-resolution degradation model, these approaches do not align with the actual physical imaging mechanism of remote sensing, rendering these algorithms impractical in real-world remote sensing applications \cite{dong2022real,zhang2024real}.

Recently, numerous deep learning-based methods have been proposed for image restoration \cite{cui2024revitalizing, qi2024representing, chen2022simple, cui2023image, zhang2018image, ren2024fast, zamir2020learning, liang2021swinir, ZHRen2024, quan2021gaussian} and have demonstrated strong performance across various datasets and evaluation metrics. These methods aimed to integrate advanced deep learning modules and techniques such as channel attention mechanisms \cite{zhang2018image}, spatial attention mechanisms \cite{zamir2020learning}, transformer mechanisms \cite{liang2021swinir}, and generative adversarial networks (GANs) \cite{wang2018esrgan}. The primary aim of these methods is to learn the mapping between LR and HR images. However, several challenges remain. 1) Most super-resolution algorithms use the first-order degradation model, which is inadequate for real images as these are often affected by high-order complex degradations; 2) the end-to-end black-box nature of deep learning-based algorithms makes it challenging to distinguish how the network architecture influences its prediction, emphasizing the need for enhanced transparency and interpretability in these frameworks; 3) many deep learning techniques focus on designing sophisticated network modules and depend on expanding them deeper and wider to enhance performance, this results in increased computational needs and challenges related to the smart design of network architectures.

To address these problems, we introduce a high-order degradation model to describe the combined effects of various complex degradations in remote sensing imaging, including atmospheric transmission degradation, sensor imaging system degradation, satellite platform flutter degradation, etc. Based on the high-order degradation imaging, we propose a new high-order progressive restoration network, called HDI-PRNet, which aims to recover multiple layers of degradation. Specifically, each stage of this model is composed of three components: a module for image denoising that relies on proximal mapping prior learning, a module for image deblurring that integrates Neumann series expansion with dual-domain degradation learning, and a module for super-resolution. The main contributions in this paper can be summarized as follows.
\begin{itemize}
\item We introduce a novel high-order progressive remote sensing restoration network, termed HDI-PRNet, constructed by a reverse Markov process. Our framework adeptly manages complex degradation mechanisms as a Markov chain, notably enhancing restoration performance, particularly in real remote sensing applications. The structures aligning reverse degradation tasks with respective modules render HDI-PRNet more transparent and interpretable.
\item A denoising module is developed using proximal mapping operator learning. By leveraging the multi-scale encoder-decoder learning and attention mechanisms, it emphasizes crucial denoising prior details required at each stage while preserving fine detail information.
\item We design a deblurring module that leverages the Neumann series expansion in combination with a dual-domain backward degradation learning block. This approach efficiently acquires prior knowledge of blur degradation operators in both spatial and frequency domains, leading to the integration of dual-domain representations.
\item Experimental results confirm that the HDI-PRNet framework can obtain high-quality restored images on various randomly degraded synthetic low-resolution images and real remote sensing images.
\end{itemize}

This paper is organized as follows. Section \ref{sec02} covers related work in image restoration. Our proposed HDI-PRNet framework is detailed in Section \ref{sec03}. we analyze our
experimental results and ablation experiments in Section \ref{sec04}. Finally, Section \ref{sec05} presents the conclusions.

\section{Related Work}\label{sec02}
In this section, we briefly review the related works on natural image restoration algorithms, remote sensing image restoration algorithms, and multiple degradation models.

\subsection{Image restoration for natural images}
Recently, with the rapid development of neural network technology, deep learning methods have achieved excellent performance in the field of computer vision. Dong et al. \cite{dong2015image} proposed SRCNN, which first applied a three-layer convolutional neural network for image super-resolution (SR). Zhang et al. \cite{zhang2018residual, zhang2020residual} proposed the residual dense block (RDB), which fully utilizes all of the layers in it through local dense connections, and proposed the residual dense network (RDN). Zhang et al. \cite{zhang2018image} proposed the structure of residuals in residuals (RIR) to form very deep networks, and also proposed the residual channel attention network (RCAN) for image SR.
Zamir et al. \cite{zamir2022restormer} introduced Restormer, a transformer model that performs well in a variety of image restoration tasks. Meanwhile, Fan et al. \cite{XHFan2023b} developed an incremental learning framework, called Nest-DGIL, using Nesterov's method to reconstruct images from compressed sensing data. Dai et al. \cite{dai2019second} proposed a high-order attention network for single image SR (SAN). Liang et al. \cite{liang2021swinir} proposed an image restoration model based on the Swin Transformer (SwinIR), and achieved good performance in image SR task, image denoising and JPEG artifact removal tasks. Mei et al. \cite{mei2021image} proposed a novel non-local sparse attention (NLSA) with dynamic sparse attention pattern for image SR.
Cui et al. \cite{cui2023image} introduced FSNet, an image restoration network focusing on frequency information selection. Chen et al. \cite{chen2022simple} introduced a nonlinear activation-free network (NAFNet), which offers a simplified and efficient baseline for recovery models by replacing the activation function with multiplication. Chen et al. \cite{chen2023dual} proposed the dual aggregation transformer (DAT), which aggregates spatial-channel features in the interblock and intra-block, achieving powerful feature representation capabilities and achieving fascinating performance on SR task. While these algorithms have demonstrated fascinating results in natural image restoration tasks, they often prioritize the development of complex high-performance learning modules and neglecting the interpretability of the network structure.

\subsection{Image restoration for remote sensing images}
Nowadays, with the increasing application of remote sensing images in various fields, the remote sensing image restoration algorithms have attracted increasing attention. They can improve the quality of remote sensing images, restoring more distinct details compared to the original low-quality images, without equipment upgrades.

With the development of deep learning and the success of natural image processing algorithms, lots of remote sensing image restoration algorithms have been proposed. Lei et al. \cite{lei2017super} designed a combined local-global network (LGCNet), which learns multilevel feature representations of images through different convolutional layers and global-local features of ground objects and environmental priors, to reconstruct high-quality remote sensing images. Zhang et al. \cite{zhang2020remote} proposed a kind of novel mixed high-order attention network (MHAN) to fully exploit hierarchical features for detail restoration. 
Xu et al. \cite{xu2022glf} designed a global-local fusion-based cloud removal algorithm (GLF-CR) that combines synthetic aperture radar and optical image feature information for remote sensing image cloud removal. Xiao et al. \cite{xiao2024ttst} designed a top-k token selective transformer (TTST) for remote sensing image super-resolution to flexibly filter information in self-attention. Wang et al. \cite{wang2023hybrid} proposed a novel U-shaped network based on hybrid attention (HAUNet), which is designed based on a multiscale structure and a hybrid convolutional attention mechanism to enhance the representation of global features. Lei et al. \cite{lei2021transformer} proposed the transformer-based enhancement network (TransENet), which utilizes the transformer block to extract the high/low-dimensional features to enhance the representation of high dimensions features. TransENet performs well on the synthetic UCMecred and AID dataset.

However, most of these image restoration algorithms use simple bicubic downsampling to simulate remote sensing image degradation, which is inconsistent with real remote sensing degradation. This is one of the reasons why the algorithms perform poorly on real remote sensing images. In order to solve this problem, we design a network based on a high-order degradation model that is closer to real degradation.

\subsection{Multiple degradation models}
The first-order degradation model (1) is widely used in image restoration algorithms and has achieved good performance. Most of these algorithms simplify degradation into a simple bicubic downsampling process. However, real-world imaging degradation is a complex, multi-faceted process, and simple bicubic downsampling limits the robustness of the algorithm and does not accurately simulate real-world image degradation. Wang et al. \cite{wang2021real} proposed a high-order degradation model that simulates complex real-world degradation by combining multiple degradations (including blur, noise, downsampling, etc.), and achieved excellent real-world super-resolution results. Zhang et al. \cite{zhang2021designing} proposed a complex degradation model which consists of multiple random shuffled degradations. Potlapalli et al. \cite{potlapalli2023promptir} designed a prompt learning restoration network (PromptIR) for different types of degradation, providing a versatile and efficient plug-in module with few lightweight prompts that can be used to restore images of various types and degradation levels. Kong et al. \cite{kong2024towards} proposed a sequential learning strategy and a prompt learning strategy for all-in-one image restoration. These two strategies have good effects on the CNN and Transformer backbones, and they can promote each other to learn effective image representations. In comparison, our work is focused on developing an unfolding network for real remote sensing image restoration that is conceptually simple but more interpretable.

\begin{figure*}[htbp]
\centering
\includegraphics[width=7in]{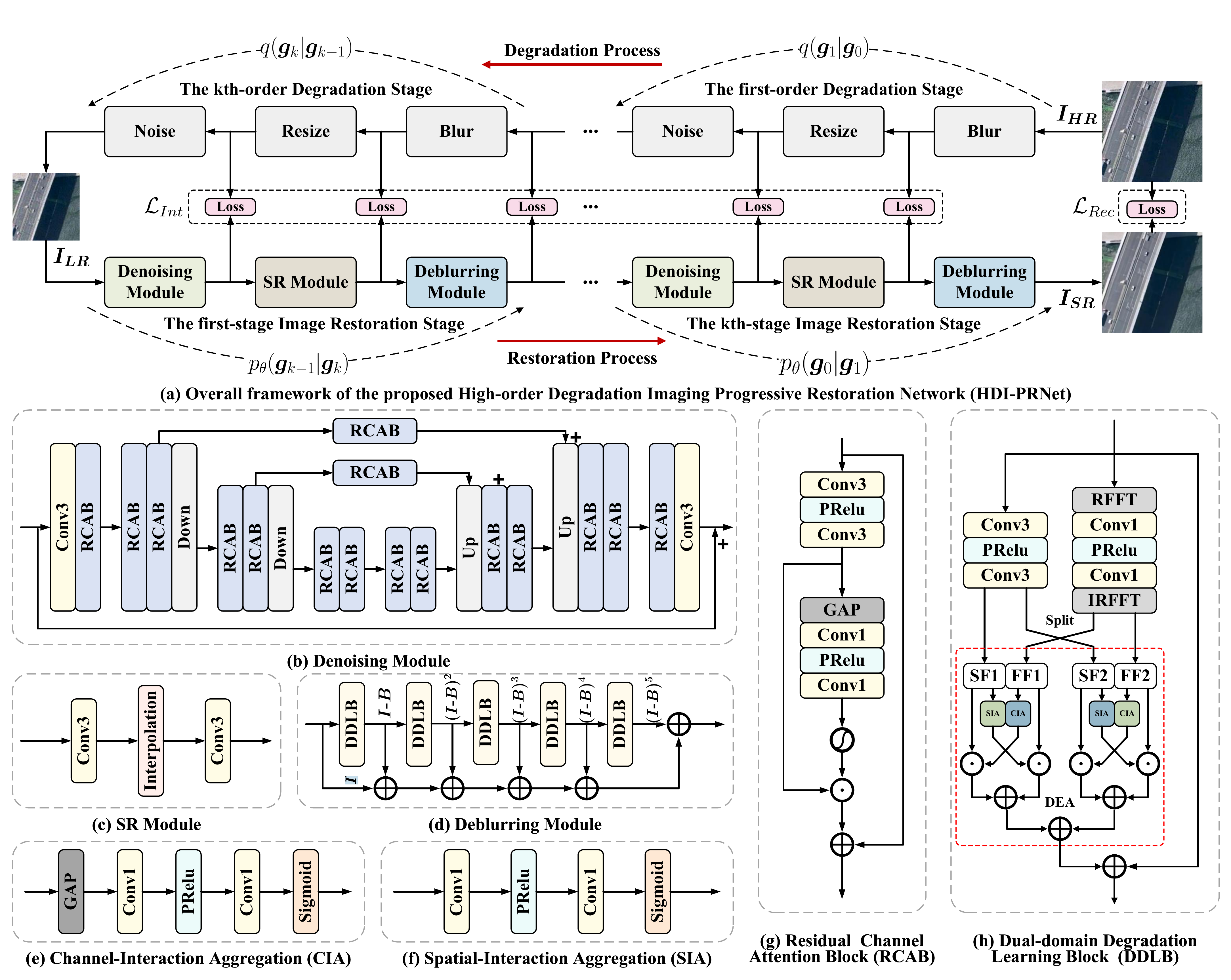}
\caption{The overall framework of the proposed HDI-PRNet solving the high-order degradation model.}
\label{fig1}
\end{figure*}

\section{Methodology}\label{sec03}
This section first briefly introduces the high-order degradation model, followed by a detailed explanation of the proposed progressive restoration architecture designed to solve high-order degradation imaging. We also introduce a multi-supervising loss function to train the proposed architecture.
\subsection{High-order Degradation Model}
An image degradation model is formulated to simulate the mechanisms that lead to a decrease in image quality \cite{gu2019blind,huang2020unfolding}. The classical first-order image degradation (\ref{eq66}) is rewritten as
\begin{equation}
\boldsymbol{g} = \mathcal{D}(\boldsymbol{u}) = (\boldsymbol{h}*\boldsymbol{u})_{\downarrow_{\boldsymbol{s}}} + \boldsymbol{n},
\end{equation}
where $\mathcal{D}$ denotes the degradation operator, it includes factors such as the blur, downsampling, and noise degradation. However, the first-order model described above remains inadequate for addressing certain complicated real-world degradation, particularly unknown noises and complex artifacts. The authors in \cite{wang2021real} improve the first-order degradation model by extending it to a more flexible higher-order degradation to better represent real-world degradations, as defined by
\begin{equation}
\boldsymbol{g} =\mathcal{D}_k(\cdots \mathcal{D}_2(\mathcal{D}_1(\boldsymbol{u}))),
\label{eq2_1}
\end{equation}
where $\mathcal{D}_\ell$ denotes the $\ell$-th imaging degradation ($1\leq\ell\leq k$).

In this work, the formula \eqref{eq2_1} can be modeled as a diffusion-based forward process. We assume $\boldsymbol{g}_0=\boldsymbol{u}$, $\boldsymbol{g}_\ell=\mathcal{D}_{\ell}(\boldsymbol{g}_{\ell-1})$ ($1\leq\ell\leq k$) and $\boldsymbol{g}_k=\boldsymbol{g}$, then decompose the $k$-order degradation framework into a series of $k$ sequential first-order models, described as follows.
\begin{equation}
\begin{split}
\boldsymbol{g}_1 =&\mathcal{D}_{1}(\boldsymbol{g}_0)=\mathcal{N}_{1}(\mathcal{S}_{1}(\mathcal{B}_{1}(\boldsymbol{g}_0))),\\
\boldsymbol{g}_2 =&\mathcal{D}_{2}(\boldsymbol{g}_1)=\mathcal{N}_{2}(\mathcal{S}_{2}(\mathcal{B}_{2}(\boldsymbol{g}_1))),\\
 &\cdots\\
\boldsymbol{g}:=&\boldsymbol{g}_k =\mathcal{D}_{k}(\boldsymbol{g}_{k-1})=\mathcal{N}_{k}(\mathcal{S}_{k}(\mathcal{B}_{k}(\boldsymbol{g}_{k-1}))),
\end{split}
\label{eq2}
\end{equation}
where $\mathcal{B}_{k}(\cdot)$, $\mathcal{S}_{k}(\cdot)$, $\mathcal{N}_{k}(\cdot)$ represent the blur, downsampling, and noise degradation operators, respectively. 

\subsection{Markov-Chain and MAP Estimation}
We formulate the higher-order degradation model \eqref{eq2_1} as a Markov chain (see Fig.\ref{fig1}(a)), so the joint probability distribution of the degraded observations $\boldsymbol{g}_{1:k}$ conditioned on the initial clean image $\boldsymbol{g}_{0}$ can be expressed as
\begin{equation}
q\left(\boldsymbol{g}_{1: k} \mid \boldsymbol{g}_{0}\right)=\prod_{\ell=1}^{k} q\left(\boldsymbol{g}_{\ell} \mid \boldsymbol{g}_{\ell-1}\right),
\label{eq2_a}
\end{equation}
where each transition $q\left(\boldsymbol{g}_{\ell} \mid \boldsymbol{g}_{\ell-1}\right)$ corresponds to a degradation step.
The formula \eqref{eq2_a} outlines a hierarchical degradation model that provides a more precise representation of the real remote sensing imaging system.

The corresponding image restoration aims to recover the original image by progressively reversing the degradation. For degraded observation $\boldsymbol{g}_k$, this reverse Markov process is modeled as
\begin{equation}
\begin{split}
p_{\theta}\left(\boldsymbol{g}_{0: k-1}|\boldsymbol{g}_k\right)=\prod_{\ell=1}^{k} p_{\theta}\left(\boldsymbol{g}_{\ell-1} \mid \boldsymbol{g}_{\ell}\right)
\end{split}
\label{eq2_b}
\end{equation}
to find the most probable recovery sequence $\boldsymbol{g}_{0: k-1}$ through a Maximum a posteriori (MAP) estimation model defined by
\begin{equation}
\begin{split}
\underset{\boldsymbol{g}_{0: k-1}}{\arg \max }\; p_{\theta}\left(\boldsymbol{g}_{0: k-1}|\boldsymbol{g}_k\right)=\underset{\boldsymbol{g}_{0: k-1}}{\arg \max } \prod_{\ell=1}^{k} p_{\theta}\left(\boldsymbol{g}_{\ell-1} \mid \boldsymbol{g}_{\ell}\right),
\end{split}
\label{eq2_c}
\end{equation}
which is progressively solved by
\begin{equation}
\bm{g}_{\ell-1}=\arg\max\limits_{\bm{g}}p_{\theta}\left(\boldsymbol{g}\mid \boldsymbol{g}_{\ell}\right), \quad \ell=k,k-1,\dots, 1.
\label{eq2_c}
\end{equation}

By applying Bayes' formula, the reverse diffusion step splits into components
\begin{equation}
p_{\theta}(\boldsymbol{g}|\boldsymbol{g}_{\ell})\propto q_{\theta}(\boldsymbol{g}_{\ell}|\boldsymbol{g})p_{\theta}(\boldsymbol{g}), \quad \ell=k,k-1,\dots, 1,
\label{eq2_d}
\end{equation}
and taking the negative log-likelihood refines the optimization (\ref{eq2_c}) as follow
\begin{equation}
\boldsymbol{g}_{\ell-1} = \underset{\boldsymbol{g}}{\arg \max } \{-\log(q_{\theta}(\boldsymbol{g}_{\ell} \mid \boldsymbol{g}))-\log(p_{\theta}(\boldsymbol{g}))\}.
\label{eq2_e}
\end{equation}

Further, we consider the following assumptions
\begin{equation}
\begin{split}
-\log(q_{\theta}(\boldsymbol{g}_{\ell} \mid \boldsymbol{g})) &\propto \frac{1}{2\sigma_\ell^2} \|\boldsymbol{g}_{\ell} -\mathcal{S}_{\ell}(\mathcal{B}_{\ell}(\boldsymbol{g}))\|,\\
-\log(p(\boldsymbol{g})) &\propto \frac{1}{2\delta_\ell^2} f_{\ell}(\boldsymbol{g}).
\end{split}
\label{eq2_f}
\end{equation}
Thus, the restoration algorithm based on Eq. \eqref{eq2_1} and Eq. \eqref{eq2} can be divided into $k$ stages to be solved progressively, where $k$ stages correspond to the high-order degradation mechanism. 
The energy minimization model (\ref{eq2_e}) at $\ell$-stage can be rewritten as
\begin{equation}
\boldsymbol{g}_{\ell-1} = \underset{\boldsymbol{g}}{\arg \min }\|\boldsymbol{g}_\ell-\mathcal{S}_{\ell}(\mathcal{B}_{\ell}(\boldsymbol{g}))\|_{2}^{2}+\lambda_{\ell} f_{\ell}(\boldsymbol{g}),
\label{eq9}
\end{equation}
where the regularization $f_{\ell}(\cdot)$ can be used to constrain the geometric prior of a given image in the denoising subproblem.

\subsection{Algorithm}
There have been several excellent explorations \cite{zhang2020deep, mou2022deep, huang2020unfolding} in solving (\ref{eq9}). One of main categories involves step-by-step degradation representations and typically consists of three components: prior denoising or local filtering, super-resolution or interpolation, and blind degradation prediction. The above three components are performed either separately (iteratively) or jointly. Therefore, the above energy minimization model \eqref{eq9} can be divided into three sub-problems, as follows
\begin{align}
\boldsymbol{g}^{dn}_{\ell} &= \underset{\bm{g}}{\arg \min }\|\bm{g} - \bm{g}_\ell\|_{2}^{2} + \lambda_{\ell} f_{\ell}(\bm{g}),\label{eq10}\\
\bm{g}^{sr}_{\ell}  &\in\{\bm{g} | \mathcal{S}_{\ell}(\bm{g})=\boldsymbol{g}^{dn}_{\ell}\},\label{eq11}\\
\bm{g}_{\ell-1}  &:= \bm{g}^{db}_{\ell}\in\{\bm{g} | \mathcal{B}_{\ell}(\bm{g})=\boldsymbol{g}^{sr}_{\ell}\}.\label{eq12}
\end{align}

The classical solution methods of the above subproblems depend on the types and levels of degradation, and the forward simulation typically involves straightforward synthetic degradations. Furthermore, any imprecise estimations of degradation in real images will unavoidably lead to artifacts. Therefore, the following literature will focus
on developing a progressive unfolding restoration network for real remote sensing imaging.

\subsection{Proposed Framework}
\subsubsection{Overall Architecture}
Deep neural networks with their powerful non-linear representation capabilities have become increasingly popular in addressing reconstruction challenges associated with degraded data. Building upon the progressive operator splitting theory described above, we construct a deep unfolding network for the degradation reconstruction. The overall framework we propose is shown in Fig. \ref{fig1}(a).

The input variable $\bm{g}_\ell$ at $\ell$-stage first undergoes a deep prior unfolding denoising sub-network to obtain the clean approximation $\boldsymbol{g}^{dn}_{\ell}$.
In addition, the super-resolution distillation SR module is proposed to exploit the structural characteristics among different features by interpolating multiscale representations, thereby improving the resolution of an image $\boldsymbol{g}^{dn}_{\ell}$ and improving image quality. This super-resolution estimation $\boldsymbol{g}^{sr}_{\ell}$ is then refined through a deep blind-deblurring subnetwork, where each block corresponds to one Neumann expansion term of the blind-deblurring sub-problem. The initial reconstruction $\boldsymbol{g}^{sr}_{\ell}$ is processed through multiple blocks of the sub-network to achieve the final deblurring reconstruction. Specifically, our innovative network design is carried out hierarchically.
The framework consists of $k$ stages corresponding to the $k$-order degradation model, and each stage consists of three modules corresponding to three sub-problems \eqref{eq10}-\eqref{eq12}.

To better leverage the prior information, we employ one proximal mapping subnetwork to independently solve (\ref{eq10}), represented as the denoising module. For the subproblem of blind deblurring, we improve the feature representation capability of the network by learning truncated Neumann series to replace the direct inverse operator $\mathcal{B}^{-1}$. This structure ensures the robustness of the framework in handling multiple or different types of degraded image. Each degradation sub-system will solve the inverse problem progressively. Additionally, to reduce information loss between network stages, we add intermediate losses to ensure that the submodule learns the corresponding subproblem, to improve the spatio-temporal feature representation of the framework, and also to ensure the interpretability of the algorithm. The specific design of each sub-network will be introduced in the following subsections.

\subsubsection{Image denoising module} The sub-problem \eqref{eq10} is a classical denoising optimization, which is highly effective in removing various types of noise. The solution of the optimization \eqref{eq10} with the general regularization term $f_{\ell}$ can be given by the proximal operator as
\begin{align}
\boldsymbol{g}^{dn}_{\ell} &= \textbf{prox}_{\lambda_{\ell}, f_{\ell}}(\bm{g}_\ell).\label{eq13}
\end{align}
Since $f_{\ell}(\cdot)$ is the geometric prior term of the solution image $\bm{g}$, hence one of the key advantages of the sub-problem \eqref{eq10} is its ability to preserve important image details. However, the regularization parameter $\lambda_\ell$ in \eqref{eq10} controls the trade-off between the fidelity to the observed data and the smoothness of the image. If $\lambda_\ell$ is larger, the denoised image is imposed stronger smoothing, leading to a potentially smoother but a loss of fine-grained details like edges, textures, and small structures. If $\lambda_\ell$ is smaller, then the denoising effect may be weaker, resulting in an image closer to the original but potentially with more noise.

The proximal mapping module can analyze the local neighborhood of each pixel and fuse features to reduce the noise component while preserving the details of the image \cite{XHFan2024a}. The optimization problem \eqref{eq10} provides a theoretical guarantee for the performance and convergence of the proximal mapping module. In other words, it typically operates by solving a proximal operator problem that seeks to minimize the denoising problem \eqref{eq10}. Compared to some complex iterative denoising methods that require a large number of calculations for each pixel and multiple iterations over the entire image, the proximal mapping module can achieve good denoising results with fewer computational resources \cite{XHFan2021a,XHFan2023b}. This makes it suitable for real-time or near-real-time applications such as video denoising or high-throughput image processing in industrial inspection.

To obtain a more accurate prior denoising expression of \eqref{eq13}, we design an image denoising unfolding module to learn the proximal mapping, as shown in Fig. 1(b). The proximal mapping $\textbf{prox}_{\lambda_{\ell}, f_{\ell}}$ is designed as a coarse to fine architecture, uses an encoder and a decoder to learn the generalized multiscale prior $\lambda_{\ell}f_{\ell}(\bm{g})$, and then obtains the solution of \eqref{eq13}.
Firstly, the denoising module adopts a convolutional design to extract shallow prior features. Secondly, it involves three scales, and we use Residual Channel Attention Blocks (RCAB) to extract and fuse multi-scale prior features at three scales (\emph{see} Fig.\ref{fig1}(b) and (g)). The skip connections between downsampling and upsampling at each scale are also processed using RCAB, and the channel number of the three scales is set to 120, 140, and 160 at resolutions $1, 1/2, 1/4$, respectively. Then, we introduce a skip connection before and after the encoder decoder to ensure the stability of proximal mapping learning. Finally, in terms of sampling operators, we use bilinear sampling to perform down-sampling and up-sampling operations.

\subsubsection{Image super-resolution module}
Image super-resolution is a method employed to improve the resolution of an image, resulting in a clearer and more detailed appearance. This enhancement is typically accomplished using algorithms and models that estimate high-resolution predictions from low-resolution inputs \cite{zhang2020deep}.

The CNN image super-resolution leverages deep learning techniques to enhance the resolution of images. It typically involves training a CNN model on low-resolution and high-resolution image pairs, where the network learns to reconstruct high-resolution images from their low-resolution counterparts. The proposed framework employs an SR module with convolution and pixelshuffle to solve the problem at $\ell$-stage as
\begin{equation}
\begin{split}
\bm{g}^{sr}_{\ell} \in\{\bm{g} | \mathcal{S}_{\ell}(\bm{g})=\boldsymbol{g}^{dn}_{\ell}\},\;\;1\leq\ell\leq k.
\end{split}
\end{equation}

Since the proposed framework is a $k$-stage cascaded architecture that processes sub-problems progressively, the first-stage SR module resizes the image from the LR space to the middle resolution space, and then the final stage resizes it to the HR space. In this case, we need to perform SR tasks with a different upsampling ratio ($\uparrow_{\boldsymbol{s}}$) in the $k$ stages. To achieve efficiently fractional super-resolution without introducing new degradation, we simply use convolution and bilinear interpolation to perform upsampling operations to solve \eqref{eq11}, as shown in Fig. 1(c). This module can ensure that the image will not have problems such as checkerboard artifacts caused by transposed convolution. However, such an SR module can efficiently adjust the spatial resolution. Although bilinear interpolation is considered to have defects in sharpening edge details, the deblurring problem can be handled by the subsequent deblurring module.

\subsubsection{Image deblurring module} Let the blurring operator $\mathcal{B}_{\ell}$ be a bounded linear operator that satisfies $\rho(\mathcal{I}-\mathcal{B}_{\ell})< 1$ on a Banach space $\mathcal{X}$. The Neumann expansion series of $\mathcal{B}_{\ell}^{-1}$ is given by the following infinite series
\begin{equation}
\begin{split}
\mathcal{B}_{\ell}^{-1}=\left(\mathcal{I}-(\mathcal{I}-\mathcal{B}_{\ell})\right)^{-1} = \sum_{i=0}^{+\infty}(\mathcal{I} - \mathcal{B}_{\ell})^{i},
\end{split}
\label{eq20}
\end{equation}
where $\mathcal{I}$ is an identity operator, $(\mathcal{I} - \mathcal{B}_{\ell})^0=\mathcal{I}$ (the identity operator) and $(\mathcal{I} - \mathcal{B}_{\ell})^{i}=(\mathcal{I} - \mathcal{B}_{\ell})\circ\cdots\circ(\mathcal{I} - \mathcal{B}_{\ell})$ ($i$ times composition of $(\mathcal{I} - \mathcal{B}_{\ell})$ with itself for $i\geq 1$). To facilitate calculations, we use a $m$-term truncated Neumann series expansion to approximate $\mathcal{B}_{\ell}^{-1}$ as follows
\begin{equation}
\begin{split}
\bm{g}_{\ell}^{db}=\mathcal{B}_{\ell}^{-1}(\bm{g}^{sr}_{\ell})&\approx \sum_{i=0}^{m}(\mathcal{I} - \mathcal{B}_{\ell})^{i}(\bm{g}^{sr}_{\ell}).
\end{split}
\label{eq21}
\end{equation}
Even if the condition $\rho(\mathcal{I} - \mathcal{B}_{\ell}) < 1$ is not met, the truncated Neumann series \eqref{eq21} can still approximate the inverse operator $\mathcal{B}_{\ell}^{-1}$, providing a form of implicit regularization \cite{quan2023neumann}.

To obtain accurately blur degradation, we use the Dual Domain Degradation Learning Block (DDLB) designed based on the ResFFT-ReLU Block \cite{mao2023intriguing} as follows
\begin{equation}
\begin{split}
\mathcal{I} - \mathcal{B}_{\ell} = \mathcal{I} + \textbf{DEA}(\mathcal{B}^{Spatial}_{\ell}, \mathcal{B}^{Frequency}_{\ell})
\end{split}
\label{eq22}
\end{equation}
to learn \eqref{eq21}, as shown in Fig. \ref{fig1}(d) and (h). Here, we split the blur operator $\mathcal{B}_{\ell}$ into a dual-domain feature representation. Therefore, DDLB includes the spatial transformation branch $\mathcal{B}^{Spatial}_{\ell}$, the frequency transformation branch $\mathcal{B}^{Frequency}_{\ell}$, the dual-domain expression aggregation (DEA) and the identity transformation $\mathcal{I}$.

In the spatial domain branch, the \textbf{Conv-PRelu-Conv} configuration is used to learn and simulate the degradation process $\mathcal{B}^{Spatial}_{\ell}$. This structure is widely adopted in CNN architectures and has been validated to effectively capture local spatial features. Regarding the branch in the frequency domain, the image is transformed into a complex tensor $\bm{z} = \bm{m}e^{j\bm{\beta}}$ by a Fourier transform, which is divided into two components: amplitude $\bm{m}$ and phase $e^{j\bm{\beta}}$. The amplitude encodes the image's color information, whereas the phase represents the texture's positional information. Similarly to the spatial domain strategy, the \textbf{Conv-PRelu-Conv} block is used to model and simulate the degradation mechanism $\mathcal{B}^{Frequency}_{\ell}$. The PRelu function helps in phase selection as it satisfies $\textbf{PRelu}(\bm{m}e^{j\bm{\beta}}) = \bm{m}\textbf{PRelu}(e^{j\bm{\beta}})$. The first convolution step provides greater flexibility in phase selection, while the last convolution consolidates the selected phases to accurately express the blur operator $\mathcal{B}^{Frequency}_{\ell}$.

The representations of degradation, $\mathcal{B}^{Spatial}_{\ell}$ in the spatial domain and $\mathcal{B}^{Frequency}_{\ell}$ in the frequency domain, are not adequately integrated when combined solely with an addition operator. To enhance the interaction and blending of spatial and frequency domain representations, we developed the dual-domain expression aggregation (DEA), which incorporates a channel-interaction aggregation (CIA) module and a spatial-interaction aggregation (SIA) module, positioned between the spatial and frequency domain branches, illustrated in Fig. \ref{fig1} (e)-(f). The application of CIA and SIA enables the adaptive synchronization of dual domain degraded features, which mutually improve their insights, refine the representation of the features and promote the integration of the degraded information.
\begin{figure*}[htbp]
\centering
\includegraphics[width=0.9\textwidth]{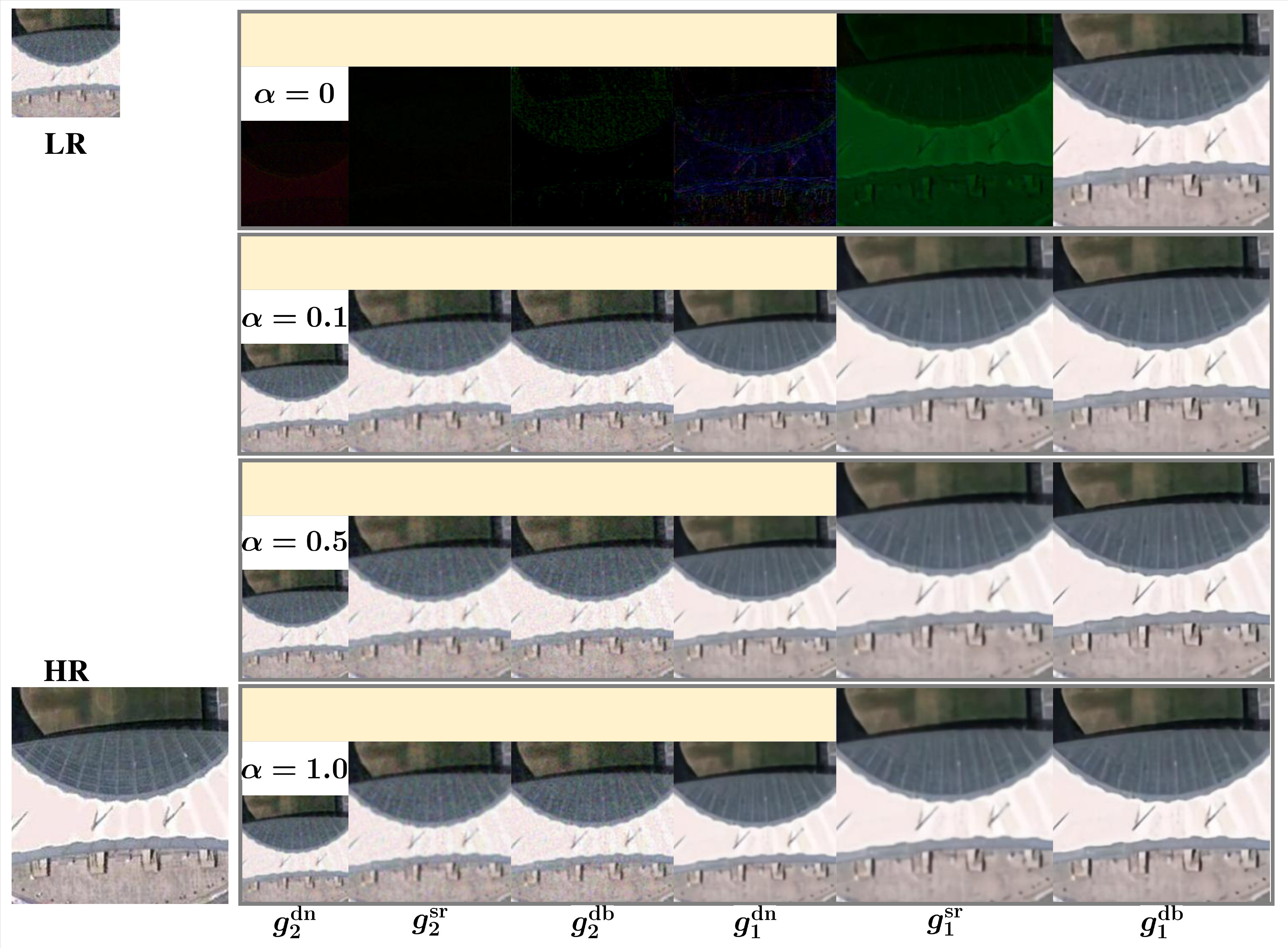}
\caption{Visualization comparison for different intermediate loss weight $\alpha$.}
\label{fig2}
\end{figure*}

\subsection{Loss Function}
The HDI-PRNet is composed of $k$ stages, with each stage containing three sub-modules. To optimize the parameters of these modules, we minimize the total loss, represented as
\begin{equation}
\begin{split}
\mathcal{L}_{Total}  = \mathcal{L}_{Rec} + \alpha \mathcal{L}_{Int},
\end{split}\label{eq24}
\end{equation}
which consists of two components: the image reconstruction loss $\mathcal{L}_{Rec}$ and the intermediate loss $\mathcal{L}_{Int}$. The weight parameter $\alpha$ is set to 1.0.

\subsubsection{Image reconstruction loss} To achieve high-quality reconstruction of the degraded image, the $\ell_2$ loss function is employed to measure the difference between the restored image $\bm{g}_{0}$ and the HR ground-truth $\boldsymbol{u}_{HR}^{gt}$. It is represented as:
\begin{equation}
\begin{split}
\mathcal{L}_{Rec}  = \|\bm{g}_{0} - \boldsymbol{u}_{HR}^{gt}\|_{2}.
\end{split}
\label{eq25}
\end{equation}

\subsubsection{Intermediate loss} In deep-supervising learning, intermediate loss assists in training by offering additional guidance in various stage modules. This loss can improve feature representation, accelerate convergence, and improve the model's predictive abilities. To verify that each module is executing its designated function, we incorporated an intermediate loss defined as follows:
\begin{equation}
\begin{split}
\mathcal{L}_{Int} &= \sum_{i\in\mathcal{A}}\|\bm{g}_{i}-\bm{u}_{i}^{gt}\|_{2},
\end{split}\label{eq26}
\end{equation}
where $\mathcal{A}$ represents the subscripts for the intermediate predictions of HDI-PRNet. In particular, the intermediate loss regulates distinct modules responsible for specific learning activities, each with a clear objective, which significantly enhances the interpretability of the proposed network.

\section{Numerical Experiments}\label{sec04}
\subsection{Implementation Details}
\subsubsection{\textbf{Datasets}}
In our experimental investigation of different remote sensing restoration methods, we have chosen to focus on well-known datasets, namely AID \cite{xia2017aid}, WHU-RS19 \cite{dai2010satellite}, and WHU Building \cite{ji2018fully}.
\begin{itemize}
\item[1)] \textbf{AID Dataset} \cite{xia2017aid} consists of 10,000 images, each measuring $600\times600$ pixels with a spatial resolution of 0.5 m/pixel (0.5 meters per pixel). This collection represents 30 different land-use categories, such as airports, farmlands, beaches, deserts, etc.
\item[2)] \textbf{WHU-RS19 Dataset} \cite{dai2010satellite} consists of 1005 remote sensing images, each $600\times600$ pixels in size, with a maximum spatial resolution of 0.5 m/pixel. The dataset represents 19 unique land types, such as airports, bridges, mountains, and rivers, among others.
\item[3)] \textbf{WHU Building Dataset} \cite{ji2018fully} features more than 220,000 individual buildings derived from aerial imagery at a spatial resolution of 0.075 m in Christchurch, New Zealand. The WHU Building dataset contains 4736 and 1036 aerial images with $512\times512$ resolution for the training and validation sets.
\item[4)] \textbf{DOTA Dataset} \cite{xia2018dota} comes from a variety of different sensors, including Google Earth, Jilin-1 satellite and Gaofen-2 satellite. The size of each image ranges from $800\times800$ to $4000\times4000$.
\item[5)] \textbf{RSSCN7 Dataset} \cite{zou2015deep} has 2,800 images, including 7 classic ground feature categories: Grass, Field, Industry, River Lake, Forest, Resident, Parking. Each category contains 400 images, and the images come from different seasons and weather conditions.
\item[6)] \textbf{UCMerced Dataset} \cite{yang2010bag} is a remote sensing image dataset containing 21 types of land objects, each category contains 100 images.
\item[7)] \textbf{NWPU-RESISC45 Dataset} \cite{cheng2017remote} contains 45 types of remote sensing scene images with a size of $256\times256$, covering a variety of scene categories such as airports, beaches, bridges, etc.
\end{itemize}
For training and assessment of our methodology, the AID and WHU Building sets are used for training, while other datasets serve as testing.

\begin{table*}
    \caption{Ablation study of model degradation orders.}
    \label{table5A}
    \renewcommand\arraystretch{1.1}
    \setlength\tabcolsep{2pt}
    \centering
    \begin{tabular*}{\hsize}{@{}@{\extracolsep{\fill}}ccccccccc@{}}
\toprule[1.5pt]
\multirow{2}{*}{{Network type}} & \multicolumn{2}{c}{{1st-order degradation images}}& \multicolumn{2}{c}{{2nd-order degradation images}} & \multicolumn{2}{c}{{3rd-order degradation images}}   & {\multirow{2}{*}{Params (M)}} & {\multirow{2}{*}{FLOPs (G)}} \\
\cmidrule(r){2-3} \cmidrule(r){4-5} \cmidrule(r){6-7}
&PSNR (dB) & SSIM  &PSNR (dB) & SSIM  &PSNR (dB) & SSIM &   &  \\
\midrule[0.95pt]
1st-order network & \textbf{31.80}  & \textbf{0.8492} & 28.81  & 0.7438 & 27.20  & 0.6776 & 8.19 & 167.37 \\
\cmidrule(r){1-9}
2nd-order network & 31.70  & 0.8467 & \textbf{30.03}  & \textbf{0.7931} & 28.49  & 0.7350 & 16.39  & 338.72 \\
\cmidrule(r){1-9}
3rd-order network & 31.48  & 0.8428 & 29.95  & 0.7913 & \textbf{28.64}  & \textbf{0.7408} & 24.58  & 515.43 \\
\cmidrule(r){1-9}
DAT & 31.32 & 0.8386 & 29.68 & 0.7855 & 28.29  & 0.7292 & 14.65  & 265.75 \\
\cmidrule(r){1-9}
TTST & 31.18 & 0.8332 & 29.45 & 0.7771 & 28.08 & 0.7203 & 18.22  & 331.30 \\
\bottomrule[1.5pt]
\end{tabular*}
\end{table*}

\subsubsection{\textbf{High-order Degradation Simulation}} Inspired by the image degradation mechanism as detailed in \cite{wang2021real}, we employ a high-order degradation model to simulate real degradation on hyper-resolution datasets. Initially, the blurring model is used on the original image, followed by applying the undersampling operator to the blurred image. Lastly, the noise model is applied to the image that has been both undersampled and blurred. We continuously implement these degradation models to simulate the $k$-order hybrid image degradation (\emph{see} Fig. \ref{fig1}(a) for more details).

To simulate the $\ell$-th blur degradation $\mathcal{B}_\ell(\cdot)$, we configure the blur kernel size as $2n+1$ where $n$ is randomly selected from the set $\{3, 4, \dots, 10\}$. We employ the selection strategy to determine the blur kernel type: 1) the sinc filter is chosen with a probability of $0.1$ to simulate the ringing and overshoot effects in images; 2) the isotropic Gaussian, anisotropic Gaussian, isotropic generalized Gaussian, anisotropic generalized Gaussian and isotropic / anisotropic plateau-shaped blur kernels are generated with a probability of $0.9$, following proportions of {0.45:0.25:0.12:0.03:0.12:0.03}. The Gaussian blur standard deviation is selected between $0.1$ and $2.0$, but for the second degradation stage, this range is narrowed to $0.1$ to $1.0$. The shape parameters for the generalized Gaussian and plateau-shaped kernels are chosen from $0.5$ to $4$ and $1$ to $2$, respectively. The second stage of blur degradation is omitted with a probability of $0.2$.

To simulate the $\ell$-th resizing degradation $\mathcal{S}_\ell(\cdot)$, we perform upsampling, downsampling, or maintain the original image resolution with probabilities of 0.2, 0.7, and 0.1, respectively (reconfigured to probabilities of 0.3, 0.4, and 0.3 during the second degradation stage). The resizing scale is randomly chosen between 0.5 and 1.5 (ranging between 0.8 and 1.2 for the second degradation step). The sampling method is randomly selected from bilinear interpolation, bicubic interpolation, and area resizing. Finally, we resize the images to achieve the desired low-resolution.

To simulate the degradation of noise pollution $\mathcal{N}_\ell(\cdot)$ at $\ell$-th stage, Gaussian and Poisson noise are incorporated, each with a probability of $0.5$. The level of Gaussian noise ranged from $1$ to $25$, and the level of Poisson noise from $0.05$ to $2.5$. In contrast, the noise level used in the second degradation stage ranges from $1$ to $20$ for Gaussian and $0.05$ to $2.0$ for Poisson. Furthermore, gray noise was applied with a probability of 0.4.

\subsubsection{\textbf{Evaluation Metrics}}
We evaluate the image restoration in remote sensing using peak signal-to-noise ratio (PSNR), structural similarity (SSIM) \cite{wang2004image}, and the blind/referenceless image spatial quality evaluator (BRISQUE) \cite{mittal2012no}. The superior restoration performance is reflected in elevated PSNR and SSIM values, along with reduced BRISQUE values.

\begin{figure*}[htbp]
\centering
\includegraphics[width=0.9\textwidth]{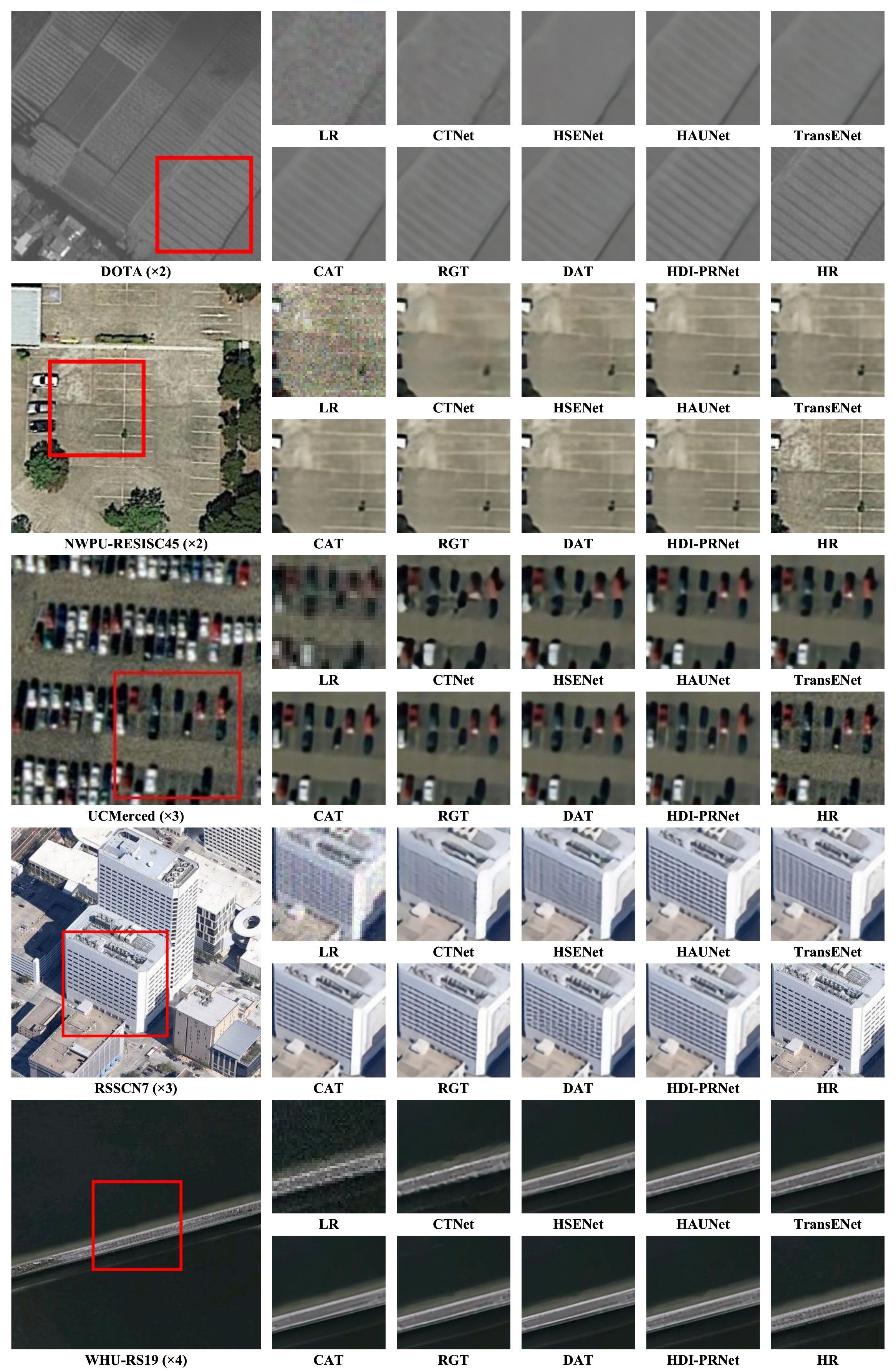}
\caption{$\times$2, $\times$3 and $\times$4 visual comparisons on DOTA, NWPU-RESISC45, UCMerced, RSSCN7 and WHU-RS19 datasets.}
\label{fig3}
\end{figure*}
\subsubsection{\textbf{Compared Methods}}
In this study, we conduct a group of studies with fourteen methods to evaluate the performance of the proposed method. Specifically, we compare our HDI-PRNet with general image restoration approaches such as RCAN \cite{zhang2018image}, SwinIR \cite{liang2021swinir}, RDN \cite{zhang2018residual}, SAN \cite{dai2019second}, DAT \cite{chen2023dual}, D-DBPN \cite{haris2018deep}, CAT \cite{chen2022cross}, and RGT \cite{chen2024recursive}, as well as super-resolution methods in remote sensing, such as MHAN \cite{zhang2020remote},  TTST \cite{xiao2024ttst}, HAUNet \cite{wang2023hybrid}, TransENet \cite{lei2021transformer}, CTNet \cite{wang2021contextual}, and HSENet \cite{lei2021hybrid}. To ensure consistency and rigor, all methods included in the study were implemented using the original code provided by the respective authors, available on their official homepages or GitHub repositories, and re-trained on the same training dataset.

\subsubsection{\textbf{Training Settings}}
We use a batch size of $32$ to train our model. Each input of the LR image is randomly cropped to dimensions of $64 \times 64$, with a total of $200K$ training iterations. The training patches undergo augmentation via random horizontal flips and rotations. To keep fair comparisons, we use the Adam optimizer with parameters $\beta_{1} = 0.9$ and $\beta_{2} = 0.99$. The initial learning rate is set as $2 \times 10^{-4}$, which is halved at all milestones of $[20K, 120K, 160K, 180K]$. Our models are implemented using PyTorch and run on a Nvidia A100 GPU.

\subsection{Ablation Study}
We conduct extensive ablation studies to evaluate the contributions of each component within the proposed HDI-PRNet framework. These experiments involve analyzing the advantages of high-order networks, the loss weight, the denoise module architecture, the number of Neumann series expansion orders, and the effectiveness of DDLB.

\subsubsection{\textbf{Evaluation for high-order degradation}}
To demonstrate the advantages of the high-order framework, we perform ablation studies, as shown in Table \ref{table5A}. Our analysis assesses the performance of frameworks with differing degradation orders across various degraded images to demonstrate their robustness. We can see from Table \ref{table5A} that the low-order degradation network exhibits significantly decreasing performance when faced with complex high-order degraded images. In contrast, a high-order degradation network achieves more consistent and superior results in different degradation images, underscoring their increased robustness. For optimal configuration, we chose the second-order degradation architecture because it provides a suitable balance between network complexity and reconstruction performance.

\subsubsection{\textbf{Sensitivity of the loss weight $\alpha$}}
The WHU-RS19 dataset is used to explore the optimal value of loss of weight $\alpha$. As shown in Table \ref{table1}, setting the weight $\alpha = 0$ results in inferior image reconstruction performance compared to the methods where $\alpha \neq 0$. This indicates that the intermediate loss that supervises each submodule is advantageous.

Fig. \ref{fig2} illustrates the intermediate predictions and the final restored results with different loss weights. Evidently, when the weight $\alpha$ is set to $0$, each sub-module predicts an approximate black-box image, obscuring the understanding of the individual roles and impacts of each sub-module. On the other hand, incorporating the intermediate loss illustrates that the sub-modules are beneficial to their designated sub-tasks. In conclusion, to preserve the clarity of the functions of each module and to enable efficient supervised learning, we have chosen $\alpha = 1.0$ as the loss weight.

\subsubsection{\textbf{Denoising module architecture}}
Table \ref{table2} presents an ablation study that demonstrates the impact of the architecture of the denoising module. We evaluate different denoising module configurations with diverse scales and blocks to determine their efficiency. According to Table \ref{table2}, the variant that employs a multi-scale architecture consistently outperforms the single-scale architecture in terms of PSNR and SSIM metrics. In addition, increasing the number of blocks also improves denoising performance. By evaluating factors such as parameters and indices, our HDI-PRNet is designed with three scales and two blocks to implement the denoising module. Compared with the baseline model with a single scale and a single block, we see a PSNR improvement of 0.1 dB.

\begin{table}
    \caption{Comparison for different loss weight parameters $\alpha$ on AIRS dataset.}
    \label{table1}
    \renewcommand\arraystretch{1.1}
    \setlength\tabcolsep{2pt}
    \centering
    \begin{tabular*}{\hsize}{@{}@{\extracolsep{\fill}}lcc@{}}
    \toprule[1.5pt]
{Loss weight $\alpha$} & {\multirow{1}{*}{PSNR (dB)}} & {\multirow{1}{*}{SSIM}} \\
\midrule[0.95pt]
 {0}              &  {29.96}            &  {0.7915}            \\
 {0.1}            &  {\bf 30.06}            &  {0.7933}            \\
 {0.5}            &  {30.04}            &  {\bf 0.7934}            \\
 {1.0}            &  {30.03}            &  {0.7931}            \\
\bottomrule[1.5pt]
    \end{tabular*}
\end{table}

\begin{table}
    \caption{{Ablation study of the denoising module configuration.}}
    \label{table2}
    \renewcommand\arraystretch{1.1}
    \setlength\tabcolsep{2pt}
    \centering
    \begin{tabular*}{\hsize}{@{}@{\extracolsep{\fill}}ccccccccc@{}}
    \toprule[1.5pt]
\multicolumn{3}{c}{{Scale number}} & \multicolumn{2}{c}{{Block number}}& {\multirow{2}{*}{Params (M)}} & {\multirow{2}{*}{FLOPs (G)}} & {\multirow{2}{*}{PSNR (dB)}}& {\multirow{2}{*}{SSIM}}\\
\cmidrule(r){1-3} \cmidrule(r){4-5}
1          & 2          & 3          & 1          & 2          & &  & &  \\
\midrule[0.95pt]
\checkmark &            &            & \checkmark &            & 7.75  & 279.94 & 29.93 & 0.7893 \\
\cmidrule(r){1-9}
           & \checkmark &            & \checkmark &            & 10.04 & 295.19 & 29.97 & 0.7904 \\
\cmidrule(r){1-9}
           &            & \checkmark & \checkmark &            & 11.98 & 298.64 & 29.98 & 0.7923 \\
\cmidrule(r){1-9}
           &            & \checkmark &            & \checkmark & 16.39 & 338.72 & {\bf 30.03} & {\bf 0.7931} \\
\bottomrule[1.5pt]
    \end{tabular*}
\end{table}
\begin{table}
    \caption{Ablation study for different orders of the Neumann
series expansion in deblurring module.}
    \label{table3}
    \renewcommand\arraystretch{1.1}
    \setlength\tabcolsep{2pt}
    \centering
    \begin{tabular*}{\hsize}{@{}@{\extracolsep{\fill}}ccccccc@{}}
    \toprule[1.5pt]
\multicolumn{3}{c}{{Orders number}} & {\multirow{2}{*}{Params (M)}} & {\multirow{2}{*}{FLOPs (G)}} & {\multirow{2}{*}{PSNR (dB)}}& {\multirow{2}{*}{SSIM}}\\
\cmidrule(r){1-3}
1          & 3          & 5          & &  &  &\\
\midrule[0.95pt]
\checkmark &            &            & 13.37 & 208.55 & 29.83 & 0.7902 \\
\cmidrule(r){1-7}
           & \checkmark &            & 14.88 & 273.64 & 29.95 & 0.7926 \\
\cmidrule(r){1-7}
           &            & \checkmark & 16.39 & 338.72 & {\bf 30.03}& {\bf 0.7931} \\
\cmidrule(r){1-7}
           & DAT &  & 14.65  & 265.75& 29.68 & 0.7855   \\
\cmidrule(r){1-7}
           & TTST &  & 18.22  & 331.30  & 29.45 & 0.7771  \\
\bottomrule[1.5pt]
    \end{tabular*}
\end{table}

\begin{table}
    \caption{Ablation study of dual-domain degradation learning block.}
    \label{table4}
    \renewcommand\arraystretch{1.1}
    \setlength\tabcolsep{2pt}
    \centering
    \begin{tabular*}{\hsize}{@{}@{\extracolsep{\fill}}ccccccc@{}}
    \toprule[1.5pt]
SB & \multicolumn{1}{c}{FB} & \multicolumn{1}{c}{DEA} & \multicolumn{1}{c}{Params (M)} & \multicolumn{1}{c}{FLOPs (G)} & \multicolumn{1}{c}{PSNR (dB)} & \multicolumn{1}{c}{SSIM} \\
\midrule[0.95pt]
\checkmark &              &  & 15.21 & 308.72 & 29.83& 0.7884 \\
\cmidrule(r){1-7}
           & \checkmark   & & 13.77 & 205.76 & 29.90  & 0.7892 \\
\cmidrule(r){1-7}
\checkmark & \checkmark   &            & 16.36 & 338.47 & 30.00  & {\bf 0.7933} \\
\cmidrule(r){1-7}
\checkmark & \checkmark   & \checkmark & 16.39 & 338.72 & {\bf 30.03}& 0.7931 \\

\bottomrule[1.5pt]
    \end{tabular*}
\end{table}

\subsubsection{\textbf{Deblurring module architecture}}
We continue our exploration by analyzing how different orders of the Neumann series expansion, specifically at orders $1$, $3$, and $5$, influence the reconstruction performance. Table \ref{table3} illustrates an improvement in performance with more expansion terms, which confirms the effectiveness of the expansion architecture. Therefore, we choose an expansion of order $5$ in the deblurring module for the optimal configuration.

It should be noted that our Neumann first-order expansion network, even in its most basic form, outperforms both DAT and TTST in reconstruction performance.

\begin{table*}
    \caption{Comparison of different methods on various remote sensing datasets. The best and second-best results are highlighted in Bold font and the underlined ones, respectively.}
    \label{table17}
    \renewcommand\arraystretch{1.1}
    \setlength\tabcolsep{2pt}
    \centering
    \begin{tabular*}{\hsize}{@{}@{\extracolsep{\fill}}lccccccccccc@{}}
    \toprule[1.5pt]
\multirow{2}{*}{Dataset} & \multirow{2}{*}{Scale} & \multicolumn{2}{c}{WHU-RS19} & \multicolumn{2}{c}{DOTA}  & \multicolumn{2}{c}{RSSCN7} & \multicolumn{2}{c}{UCMerced} & \multicolumn{2}{c}{NWPU-RESISC45} \\
                                                    \cmidrule(r){3-4} \cmidrule(r){5-6} \cmidrule(r){7-8} \cmidrule(r){9-10} \cmidrule(r){11-12}
& & {PSNR} & {SSIM} & {PSNR} & {SSIM} & {PSNR} & {SSIM} & {PSNR} & {SSIM} & {PSNR} & {SSIM} \\
\midrule[0.95pt]
{CTNet \cite{wang2021contextual}}       & {$\times 2$}   & {28.64} & {0.7569} & {30.09} & {0.7876} & {27.39} & {0.6799} & {26.43} & {0.7048} & {27.02} & {0.6975} \\
{HSENet \cite{lei2021hybrid}}           & {$\times 2$}   & {28.94} & {0.7616} & {30.70} & {0.7964} & {27.71} & {0.6853} & {26.79} & {0.7114} & {27.40} & {0.7033} \\
{D-DBPN \cite{haris2018deep}}           & {$\times 2$}   & {29.20} & {0.7698} & {30.41} & {0.7987} & {27.81} & {0.6920} & {26.99} & {0.7231} & {27.52} & {0.7122} \\
{RDN \cite{zhang2018residual}}          & {$\times 2$}   & {29.33} & {0.7740} & {30.85} & {0.8053} & {27.88} & {0.6956} & {27.06} & {0.7290} & {27.63} & {0.7159} \\
{TransENet  \cite{lei2021transformer}}  & {$\times 2$}   & {29.33} & {0.7756} & {30.92} & {0.8079} & {27.89} & {0.6977} & {27.13} & {0.7308} & {27.66} & {0.7177} \\
{RCAN \cite{zhang2018image}}            & {$\times 2$}   & {29.46} & {0.7786} & {31.03} & {0.8113} & {27.98} & {0.7016} & {27.23} & {0.7350} & {27.72} & {0.7214} \\
{SAN \cite{dai2019second}}              & {$\times 2$}   & {29.38} & {0.7742} & {30.99} & {0.8075} & {27.94} & {0.6970} & {27.14} & {0.7290} & {27.68} & {0.7169} \\
{MHAN  \cite{zhang2020remote}}          & {$\times 2$}   & {29.40} & {0.7761} & {30.87} & {0.8076} & {27.92} & {0.6976} & {27.14} & {0.7305} & {27.68} & {0.7184} \\
{SwinIR \cite{liang2021swinir}}         & {$\times 2$}   & {29.41} & {0.7778} & {30.97} & {0.8100} & {27.93} & {0.7008} & {27.16} & {0.7352} & {27.68} & {0.7206} \\
{HAUNet \cite{wang2023hybrid}}          & {$\times 2$}   & {29.52} & {0.7789} & \underline{31.25} & {0.8131} & {28.01} & {0.7003} & {27.37} & {0.7399} & {27.78} & {0.7205} \\
{TTST  \cite{xiao2024ttst}}             & {$\times 2$}   & {29.45} & {0.7771} & {31.09} & {0.8118} & {27.97} & {0.7008} & {27.25} & {0.7356} & {27.73} & {0.7200} \\
{CAT \cite{chen2022cross}}              & {$\times 2$}   & {29.63} & {0.7838} & {31.12} & {0.8152} & {28.05} & {0.7053} & {27.39} & {0.7427} & {27.82} & {0.7253} \\
{RGT \cite{chen2024recursive}}          & {$\times 2$}   & {29.63} & {0.7840} & {31.17} & {0.8156} & {28.09} & {0.7058} & {27.39} & {0.7430} & {27.86} & {0.7263} \\
{DAT \cite{chen2023dual}}               & {$\times 2$}   & \underline{29.68} & \underline{0.7855} & {31.22} & \underline{0.8166} & \underline{28.11} & \underline{0.7074} & \underline{27.42} & \underline{0.7448} & \underline{27.89} & \underline{0.7276} \\
{HDI-PRNet}                             & {$\times 2$}   & \textbf{30.03} & \textbf{0.7931} & \textbf{31.66} & \textbf{0.8230} & \textbf{28.32} & \textbf{0.7135} & \textbf{27.82} & \textbf{0.7542} & \textbf{28.13} & \textbf{0.7334} \\
\midrule[0.95pt]
{CTNet \cite{wang2021contextual}}       & {$\times 3$}   & {27.59} & {0.7200} & {29.39} & {0.7603} & {26.72} & {0.6428} & {25.61} & {0.6652} & {26.32} & {0.6579} \\
{HSENet \cite{lei2021hybrid}}           & {$\times 3$}   & {27.98} & {0.7292} & {29.74} & {0.7691} & {26.98} & {0.6500} & {25.86} & {0.6756} & {26.50} & {0.6656} \\
{D-DBPN \cite{haris2018deep}}           & {$\times 3$}   & {28.31} & {0.7431} & {29.76} & {0.7767} & {27.20} & {0.6616} & {26.26} & {0.6927} & {26.79} & {0.6798} \\
{RDN \cite{zhang2018residual}}          & {$\times 3$}   & {28.38} & {0.7449} & {30.05} & {0.7807} & {27.21} & {0.6622} & {26.20} & {0.6938} & {26.79} & {0.6803} \\
{TransENet  \cite{lei2021transformer}}  & {$\times 3$}   & {28.44} & {0.7482} & {30.05} & {0.7828} & {27.24} & {0.6642} & {26.27} & {0.6971} & {26.81} & {0.6818} \\
{RCAN \cite{zhang2018image}}            & {$\times 3$}   & {28.48} & {0.7513} & {30.17} & {0.7865} & {27.27} & {0.6681} & {26.29} & {0.6991} & {26.85} & {0.6852} \\
{SAN \cite{dai2019second}}              & {$\times 3$}   & {28.31} & {0.7435} & {30.03} & {0.7801} & {27.19} & {0.6616} & {26.19} & {0.6927} & {26.76} & {0.6789} \\
{MHAN  \cite{zhang2020remote}}          & {$\times 3$}   & {28.45} & {0.7485} & {30.10} & {0.7836} & {27.26} & {0.6654} & {26.30} & {0.6986} & {26.87} & {0.6842} \\
{SwinIR \cite{liang2021swinir}}         & {$\times 3$}   & {28.53} & {0.7523} & {30.15} & {0.7855} & {27.29} & {0.6686} & {26.31} & {0.7015} & {26.88} & {0.6854} \\
{HAUNet \cite{wang2023hybrid}}          & {$\times 3$}   & {28.51} & {0.7483} & {30.25} & {0.7855} & {27.31} & {0.6649} & {26.44} & {0.7045} & {26.96} & {0.6861} \\
{TTST  \cite{xiao2024ttst}}             & {$\times 3$}   & {28.51} & {0.7528} & {30.32} & {0.7887} & {27.32} & {0.6693} & {26.41} & {0.7036} & {26.90} & {0.6864} \\
{CAT \cite{chen2022cross}}              & {$\times 3$}   & {28.67} & {0.7573} & \underline{30.39} & {0.7918} & {27.38} & {0.6731} & {26.45} & {0.7084} & {26.97} & {0.6904} \\
{RGT \cite{chen2024recursive}}          & {$\times 3$}   & {28.70} & {0.7580} & \underline{30.39} & {0.7919} & \underline{27.42} & {0.6742} & {26.47} & {0.7083} & {27.00} & {0.6908} \\
{DAT \cite{chen2023dual}}               & {$\times 3$}   & \underline{28.72} & \underline{0.7587} & {30.37} & \underline{0.7925} & {27.41} & \underline{0.6748} & \underline{26.48} & \underline{0.7088} & \underline{27.01} & \underline{0.6918} \\
{HDI-PRNet}                             & {$\times 3$}   & \textbf{28.96} & \textbf{0.7623} & \textbf{30.72} & \textbf{0.7956} & \textbf{27.57} & \textbf{0.6766} & \textbf{26.85} & \textbf{0.7157} & \textbf{27.22} & \textbf{0.6953} \\
\midrule[0.95pt]
{CTNet \cite{wang2021contextual}}       & {$\times 4$}   & {26.74} & {0.6707} & {28.68} & {0.7274} & {26.20} & {0.6040} & {24.78} & {0.6175} & {25.68} & {0.6182} \\
{HSENet \cite{lei2021hybrid}}           & {$\times 4$}   & {27.49} & {0.7066} & {29.29} & {0.7531} & {26.64} & {0.6323} & {25.46} & {0.6560} & {26.15} & {0.6465} \\
{D-DBPN \cite{haris2018deep}}           & {$\times 4$}   & {27.42} & {0.7049} & {29.23} & {0.7508} & {26.63} & {0.6311} & {25.46} & {0.6528} & {26.15} & {0.6457} \\
{RDN \cite{zhang2018residual}}          & {$\times 4$}   & {27.42} & {0.7050} & {29.29} & {0.7503} & {26.61} & {0.6308} & {25.41} & {0.6537} & {26.15} & {0.6455} \\
{TransENet  \cite{lei2021transformer}}  & {$\times 4$}   & {27.53} & {0.7102} & {29.40} & {0.7568} & {26.66} & {0.6354} & {25.54} & {0.6612} & {26.20} & {0.6500} \\
{RCAN \cite{zhang2018image}}            & {$\times 4$}   & {27.55} & {0.7127} & {29.34} & {0.7570} & {26.68} & {0.6372} & {25.52} & {0.6629} & {26.20} & {0.6523} \\
{SAN \cite{dai2019second}}              & {$\times 4$}   & {27.38} & {0.7024} & {29.28} & {0.7494} & {26.58} & {0.6291} & {25.36} & {0.6510} & {26.09} & {0.6424} \\
{MHAN  \cite{zhang2020remote}}          & {$\times 4$}   & {27.48} & {0.7071} & {29.38} & {0.7543} & {26.63} & {0.6327} & {25.44} & {0.6566} & {26.17} & {0.6475} \\
{SwinIR \cite{liang2021swinir}}         & {$\times 4$}   & {27.65} & {0.7152} & {29.39} & {0.7579} & {26.72} & {0.6391} & {25.60} & {0.6657} & {26.25} & {0.6533} \\
{HAUNet \cite{wang2023hybrid}}          & {$\times 4$}   & {27.59} & {0.7117} & {29.47} & {0.7586} & {26.71} & {0.6362} & {25.68} & {0.6669} & {26.27} & {0.6521} \\
{TTST  \cite{xiao2024ttst}}             & {$\times 4$}   & {27.61} & {0.7138} & {29.40} & {0.7587} & {26.70} & {0.6376} & {25.57} & {0.6639} & {26.23} & {0.6521} \\
{CAT \cite{chen2022cross}}              & {$\times 4$}   & {27.70} & {0.7176} & {29.56} & {0.7618} & {26.76} & {0.6407} & {25.67} & {0.6692} & {26.28} & {0.6553} \\
{RGT \cite{chen2024recursive}}          & {$\times 4$}   & \underline{27.79} & {0.7199} & \underline{29.59} & \underline{0.7644} & \underline{26.80} & {0.6430} & {25.72} & \underline{0.6729} & {26.33} & \underline{0.6579} \\
{DAT \cite{chen2023dual}}               & {$\times 4$}   & \underline{27.79} & \underline{0.7203} & \underline{29.59} & {0.7639} & \underline{26.80} & \underline{0.6431} & \underline{25.75} & {0.6721} & \underline{26.35} & \underline{0.6579} \\
{HDI-PRNet}                             & {$\times 4$}   & \textbf{27.87} & \textbf{0.7214} & \textbf{29.78} & \textbf{0.7655} & \textbf{26.91} & \textbf{0.6434} & \textbf{25.92} & \textbf{0.6732} & \textbf{26.46} & \textbf{0.6586} \\
\bottomrule[1.5pt]
    \end{tabular*}
\end{table*}
\subsubsection{\textbf{DDLB architecture}}
We evaluate the performance of DDLB by carrying out comparative analyses with several alternative variants. Table \ref{table4} displays the PSNR and SSIM metrics with a super-resolution scaling factor of $\times 2$. It is observed that focusing solely on either the spatial or frequency branch presents difficulties in efficiently recovering the image from blur degradation. The DEA module integrates these two branches more effectively than a straightforward addition. Compared to the baseline module in the spatial branch, the PSNR achieves an improvement of 0.2dB.

\begin{table*}
    \caption{Comparison of different SR $\times$2 models on various scene classes of WHU-RS19 (PSNR ($dB$)). The best and second-best results are highlighted in Bold font and underlined ones, respectively.}
    \label{table18}
    \renewcommand\arraystretch{1.1}
    \setlength\tabcolsep{3.5pt}
    \centering
    \begin{tabular*}{\hsize}{@{}@{\extracolsep{\fill}}l*{15\centering}{p{5mm}}@{}}
    \toprule[1.5pt]
\multirow{2}{*}{Class} & \multirow{1}{*}{CTNet} & \multirow{1}{*}{HSENet} & \multirow{1}{*}{D-DBPN} & \multirow{1}{*}{RDN} & \multirow{1}{*}{TransE-} & \multirow{1}{*}{RCAN} & \multirow{1}{*}{SAN} & \multirow{1}{*}{MHAN} & \multirow{1}{*}{SwinIR} & \multirow{1}{*}{HAUNet} & \multirow{1}{*}{TTST} & \multirow{1}{*}{CAT} & \multirow{1}{*}{RGT} & \multirow{1}{*}{DAT} & \multirow{1}{*}{HDI-}   \\
& \multirow{1}{*}{\cite{wang2021contextual}} & \multirow{1}{*}{\cite{lei2021hybrid}} & \multirow{1}{*}{\cite{haris2018deep}} & \multirow{1}{*}{\cite{zhang2018residual}} & \multirow{1}{*}{Net\cite{lei2021transformer}} & \multirow{1}{*}{\cite{zhang2018image}} & \multirow{1}{*}{\cite{dai2019second}} & \multirow{1}{*}{\cite{zhang2020remote}} & \multirow{1}{*}{\cite{liang2021swinir}} & \multirow{1}{*}{\cite{wang2023hybrid}} & \multirow{1}{*}{\cite{xiao2024ttst}} & \multirow{1}{*}{\cite{chen2022cross}} & \multirow{1}{*}{\cite{chen2024recursive}} & \multirow{1}{*}{\cite{chen2023dual}} & \multirow{1}{*}{PRNet}   \\
\midrule[0.95pt]
{Airport}                          & {27.08} & {27.55} & {27.80} & {27.90} & {27.92} & {27.97} & {27.98} & {27.91} & {27.95} & {28.05} & {27.97} & {28.10} & {28.19} & \underline{28.25} & \textbf{28.55} \\
{Beach}                            & {34.41} & {34.40} & {35.67} & {36.18} & {36.28} & {36.73} & {36.64} & {36.49} & {37.11} & {36.60} & {36.55} & {37.15} & \underline{37.18} & {36.99} & \textbf{37.80} \\
{Bridge}                           & {31.79} & {32.25} & {32.54} & {32.65} & {32.54} & {32.83} & {32.49} & {32.58} & {32.58} & {32.81} & {32.83} & \underline{32.93} & {32.83} & {32.83} & \textbf{33.48} \\
{Commercial}                       & {24.31} & {24.60} & {24.74} & {24.79} & {24.84} & {24.94} & {24.87} & {24.88} & {24.88} & {24.97} & {24.94} & \underline{25.11} & {25.09} & \underline{25.11} & \textbf{25.44} \\
{Desert}                           & {36.79} & {36.83} & {37.03} & {37.29} & {37.33} & {37.43} & {37.22} & {37.38} & {37.46} & {37.50} & {37.38} & {37.53} & \underline{37.63} & \textbf{37.65} & {37.58} \\
{Farmland}                         & {33.52} & {33.95} & {33.82} & {34.16} & {33.99} & {34.28} & {34.27} & {34.23} & {34.23} & {34.40} & {34.33} & {34.36} & {34.50} & \underline{34.53} & \textbf{34.59} \\
{Football Field}                   & {28.05} & {28.39} & {28.76} & {28.89} & {29.09} & {29.14} & {28.97} & {28.96} & {29.00} & {29.24} & {29.01} & {29.42} & {29.37} & \underline{29.48} & \textbf{29.73} \\
{Forest}                           & {27.03} & {27.18} & {27.39} & {27.44} & {27.47} & {27.41} & {27.45} & {27.43} & {27.36} & {27.50} & {27.44} & \underline{27.57} & {27.50} & {27.52} & \textbf{27.84} \\
{Industrial}                       & {25.54} & {25.94} & {26.12} & {26.22} & {26.37} & {26.41} & {26.29} & {26.37} & {26.33} & {26.53} & {26.36} & {26.56} & {26.56} & \underline{26.60} & \textbf{27.08} \\
{Meadow}                           & {34.81} & {34.84} & {34.95} & {35.06} & {34.96} & {35.15} & {35.12} & {35.02} & {35.13} & {35.17} & {35.13} & {35.15} & \underline{35.21} & {35.20} & \textbf{35.22} \\
{Mountain}                         & {24.46} & {24.66} & {24.81} & {24.83} & {24.87} & {24.92} & {24.87} & {24.89} & {24.88} & {25.00} & {24.89} & {24.95} & {24.95} & \underline{25.08} & \textbf{25.27} \\
{Park}                             & {27.44} & {27.84} & {27.91} & {27.99} & {28.03} & {28.09} & {28.05} & {28.07} & {28.04} & {28.10} & {28.05} & {28.18} & \underline{28.25} & \underline{28.25} & \textbf{28.48} \\
{Parking}                          & {25.81} & {26.52} & {26.92} & {26.93} & {26.94} & {27.01} & {26.92} & {27.14} & {26.94} & {27.31} & {27.18} & \underline{27.46} & {27.27} & {27.39} & \textbf{28.18} \\
{Pond}                             & {35.10} & {35.57} & {35.56} & {35.82} & {35.65} & {35.97} & {35.82} & {35.81} & {35.89} & {36.03} & {35.94} & {35.99} & {36.02} & \underline{36.16} & \textbf{36.24} \\
{Port}                             & {26.62} & {27.03} & {27.31} & {27.31} & {27.33} & {27.39} & {27.35} & {27.48} & {27.40} & {27.43} & {27.48} & {27.70} & {27.67} & \underline{27.76} & \textbf{28.35} \\
{Railway Station}                  & {24.56} & {24.81} & {25.12} & {25.19} & {25.08} & {25.21} & {25.17} & {25.19} & {25.02} & {25.29} & {25.29} & {25.38} & {25.31} & \underline{25.50} & \textbf{26.05} \\
{Residential}                      & {24.20} & {24.48} & {24.74} & {24.74} & {24.82} & {24.90} & {24.83} & {24.89} & {24.86} & {24.90} & {24.93} & {25.10} & {25.10} & \underline{25.16} & \textbf{25.58} \\
{River}                            & {27.63} & {27.77} & {27.94} & {28.01} & {27.98} & {28.02} & {28.09} & {28.10} & {27.94} & {28.09} & {28.12} & {28.18} & {28.15} & \underline{28.22} & \textbf{28.54} \\
{Viaduct}                          & {25.17} & {25.45} & {25.96} & {26.04} & {26.09} & {26.24} & {26.03} & {26.13} & {26.12} & {26.25} & {26.10} & {26.46} & {26.46} & \underline{26.53} & \textbf{26.87} \\
\bottomrule[1.5pt]
    \end{tabular*}
\end{table*}

\begin{table*}
    \caption{Comparison of different SR $\times$2 models on various scene classes of RSSCN7 (PSNR ($dB$)). The best and second-best results are highlighted in Bold font and underlined ones, respectively.}
    \label{table19}
    \renewcommand\arraystretch{1.1}
    \setlength\tabcolsep{3.5pt}
    \centering
    \begin{tabular*}{\hsize}{@{}@{\extracolsep{\fill}}l*{15\centering}{p{5mm}}@{}}
    \toprule[1.5pt]
\multirow{2}{*}{Class} & \multirow{1}{*}{CTNet} & \multirow{1}{*}{HSENet} & \multirow{1}{*}{D-DBPN} & \multirow{1}{*}{RDN} & \multirow{1}{*}{TransE-} & \multirow{1}{*}{RCAN} & \multirow{1}{*}{SAN} & \multirow{1}{*}{MHAN} & \multirow{1}{*}{SwinIR} & \multirow{1}{*}{HAUNet} & \multirow{1}{*}{TTST} & \multirow{1}{*}{CAT} & \multirow{1}{*}{RGT} & \multirow{1}{*}{DAT} & \multirow{1}{*}{HDI-}   \\
& \multirow{1}{*}{\cite{wang2021contextual}} & \multirow{1}{*}{\cite{lei2021hybrid}} & \multirow{1}{*}{\cite{haris2018deep}} & \multirow{1}{*}{\cite{zhang2018residual}} & \multirow{1}{*}{Net\cite{lei2021transformer}} & \multirow{1}{*}{\cite{zhang2018image}} & \multirow{1}{*}{\cite{dai2019second}} & \multirow{1}{*}{\cite{zhang2020remote}} & \multirow{1}{*}{\cite{liang2021swinir}} & \multirow{1}{*}{\cite{wang2023hybrid}} & \multirow{1}{*}{\cite{xiao2024ttst}} & \multirow{1}{*}{\cite{chen2022cross}} & \multirow{1}{*}{\cite{chen2024recursive}} & \multirow{1}{*}{\cite{chen2023dual}} & \multirow{1}{*}{PRNet}   \\
\midrule[0.95pt]
{Grass}                    & {31.81} & {32.12} & {32.13} & {32.27} & {32.24} & {32.37} & {32.31} & {32.28} & {32.32} & {32.37} & {32.36} & {32.40} & {32.46} & \underline{32.48} & \textbf{32.55}  \\
{Field}                    & {31.38} & {31.63} & {31.57} & {31.75} & {31.74} & {31.86} & {31.82} & {31.78} & {31.82} & {31.90} & {31.87} & {31.91} & {31.98} & \underline{32.01} & \textbf{32.09}  \\
{Industry}                 & {24.41} & {24.85} & {25.03} & {25.08} & {25.10} & {25.17} & {25.16} & {25.13} & {25.11} & {25.24} & {25.17} & {25.28} & {25.32} & \underline{25.37} & \textbf{25.75}  \\
{River Lake}               & {29.24} & {29.60} & {29.74} & {29.79} & {29.79} & {29.86} & {29.83} & {29.84} & {29.84} & {29.86} & {29.85} & {29.94} & {29.97} & \underline{29.99} & \textbf{30.17}  \\
{Forest}                   & {27.02} & {27.24} & {27.31} & {27.36} & {27.38} & {27.43} & {27.39} & {27.38} & {27.41} & {27.43} & {27.42} & {27.49} & {27.50} & \underline{27.52} & \textbf{27.61}  \\
{Resident}                 & {23.67} & {24.06} & {24.27} & {24.28} & {24.35} & {24.41} & {24.35} & {24.35} & {24.36} & {24.47} & {24.37} & {24.53} & {24.51} & \underline{24.54} & \textbf{24.87}  \\
{Parking}                  & {24.18} & {24.44} & {24.61} & {24.62} & {24.65} & {24.72} & {24.69} & {24.68} & {24.67} & {24.76} & {24.73} & {24.84} & {24.87} & \underline{24.89} & \textbf{25.17}  \\
\bottomrule[1.5pt]
    \end{tabular*}
\end{table*}

\begin{table*}
    \caption{Comparison of different SR $\times$4 models on various scene classes of UCMerced (PSNR ($dB$)). The best and second-best results are highlighted in Bold font and underlined ones, respectively.}
    \label{table20}
    \renewcommand\arraystretch{1.1}
    \setlength\tabcolsep{3.5pt}
    \centering
    \begin{tabular*}{\hsize}{@{}@{\extracolsep{\fill}}l*{15\centering}{p{5mm}}@{}}
    \toprule[1.5pt]
\multirow{2}{*}{Class} & \multirow{1}{*}{CTNet} & \multirow{1}{*}{HSENet} & \multirow{1}{*}{D-DBPN} & \multirow{1}{*}{RDN} & \multirow{1}{*}{TransE-} & \multirow{1}{*}{RCAN} & \multirow{1}{*}{SAN} & \multirow{1}{*}{MHAN} & \multirow{1}{*}{SwinIR} & \multirow{1}{*}{HAUNet} & \multirow{1}{*}{TTST} & \multirow{1}{*}{CAT} & \multirow{1}{*}{RGT} & \multirow{1}{*}{DAT} & \multirow{1}{*}{HDI-}   \\
& \multirow{1}{*}{\cite{wang2021contextual}} & \multirow{1}{*}{\cite{lei2021hybrid}} & \multirow{1}{*}{\cite{haris2018deep}} & \multirow{1}{*}{\cite{zhang2018residual}} & \multirow{1}{*}{Net\cite{lei2021transformer}} & \multirow{1}{*}{\cite{zhang2018image}} & \multirow{1}{*}{\cite{dai2019second}} & \multirow{1}{*}{\cite{zhang2020remote}} & \multirow{1}{*}{\cite{liang2021swinir}} & \multirow{1}{*}{\cite{wang2023hybrid}} & \multirow{1}{*}{\cite{xiao2024ttst}} & \multirow{1}{*}{\cite{chen2022cross}} & \multirow{1}{*}{\cite{chen2024recursive}} & \multirow{1}{*}{\cite{chen2023dual}} & \multirow{1}{*}{PRNet}   \\
\midrule[0.95pt]
{Agricultural}                & {25.07} & {25.33} & {25.28} & {25.36} & {25.55} & {25.52} & {25.35} & {25.41} & {25.60} & {25.63} & {25.59} & {25.67} & \textbf{25.73} & {25.70} & \underline{25.72} \\
{Airplane}                    & {24.91} & {25.76} & {25.74} & {25.63} & {25.76} & {25.86} & {25.64} & {25.75} & {25.91} & {25.94} & {25.86} & {25.96} & \underline{26.03} & {26.02} & \textbf{26.33} \\
{Baseball Diamond}             & {29.15} & {29.71} & {29.72} & {29.74} & {29.71} & {29.74} & {29.65} & {29.73} & {29.87} & \underline{29.93} & {29.86} & {29.87} & {29.88} & \underline{29.93} & \textbf{30.13} \\
{Beach}                       & {31.22} & {31.66} & {31.50} & {31.51} & {31.69} & {31.59} & {31.57} & {31.57} & {31.70} & {31.77} & {31.71} & {31.74} & {31.79} & \textbf{31.82} & \underline{31.80} \\
{Buildings}                   & {22.79} & {23.68} & {23.75} & {23.57} & {23.74} & {23.80} & {23.54} & {23.60} & {23.82} & {23.88} & {23.83} & {23.97} & {23.99} & \underline{24.08} & \textbf{24.24} \\
{Chaparral}                   & {22.93} & {23.73} & {23.65} & {23.72} & {23.94} & {23.76} & {23.61} & {23.71} & {23.93} & \textbf{23.97} & \underline{23.95} & {23.91} & {23.79} & {23.89} & {23.62} \\
{Dense Residential}            & {22.97} & {23.67} & {23.70} & {23.57} & {23.77} & {23.71} & {23.55} & {23.61} & {23.80} & {23.91} & {23.77} & {23.90} & \underline{23.98} & {23.97} & \textbf{24.30} \\
{Forest}                      & {25.40} & {25.73} & {25.73} & {25.72} & {25.79} & {25.78} & {25.68} & {25.74} & {25.82} & {25.84} & {25.80} & {25.83} & \underline{25.87} & {25.86} & \textbf{25.88} \\
{Freeway}                     & {25.23} & {26.43} & {26.46} & {26.44} & {26.47} & {26.63} & {26.30} & {26.48} & {26.52} & {26.67} & {26.52} & {26.69} & {26.79} & \underline{26.82} & \textbf{27.01} \\
{Golfcourse}                  & {29.33} & {29.81} & {29.76} & {29.81} & {29.92} & {29.83} & {29.78} & {29.83} & {29.90} & {29.93} & {29.83} & {29.93} & {29.99} & \underline{30.00} & \textbf{30.12} \\
{Harbor}                      & {20.49} & {21.26} & {21.32} & {21.14} & {21.39} & {21.25} & {21.12} & {21.18} & {21.44} & {21.65} & {21.34} & {21.59} & {21.57} & \underline{21.67} & \textbf{22.05} \\
{Intersection}                & {24.04} & {24.72} & {24.78} & {24.66} & {24.77} & {24.84} & {24.67} & {24.73} & {24.84} & {24.91} & {24.86} & {24.96} & {25.01} & \underline{25.03} & \textbf{25.27} \\
{Medium Residential}           & {23.31} & {23.88} & {23.90} & {23.84} & {23.96} & {23.93} & {23.81} & {23.87} & {24.02} & {24.07} & {23.98} & {24.08} & {24.13} & \underline{24.19} & \textbf{24.29} \\
{Mobile Home Park}              & {21.50} & {22.17} & {22.21} & {22.03} & {22.14} & {22.13} & {22.09} & {22.06} & {22.28} & {22.35} & {22.19} & {22.39} & \underline{22.45} & {22.43} & \textbf{22.71} \\
{Overpass}                    & {23.88} & {24.82} & {24.90} & {24.86} & {24.89} & {25.02} & {24.69} & {24.89} & {25.02} & {25.04} & {24.95} & {25.11} & {25.25} & \underline{25.27} & \textbf{25.46} \\
{Parking Lot}                  & {20.02} & {20.71} & {20.76} & {20.57} & {20.94} & {20.78} & {20.57} & {20.65} & {20.98} & {21.11} & {20.85} & {21.13} & {21.16} & \underline{21.23} & \textbf{21.49} \\
{River}                       & {25.83} & {26.16} & {26.18} & {26.16} & {26.24} & {26.18} & {26.13} & {26.20} & {26.24} & \underline{26.30} & {26.22} & {26.24} & {26.26} & {26.28} & \textbf{26.32} \\
{Runway}                      & {26.72} & {27.87} & {27.70} & {27.63} & {27.86} & {27.81} & {27.65} & {27.77} & {27.87} & {28.09} & {28.02} & {27.96} & \underline{28.14} & {28.13} & \textbf{28.59} \\
{Sparse Residential}           & {25.27} & {25.81} & {25.82} & {25.81} & {25.87} & {25.87} & {25.71} & {25.82} & {25.97} & {25.99} & {25.91} & {25.99} & {26.01} & \underline{26.04} & \textbf{26.13} \\
{Storage Tanks}                & {24.91} & {25.63} & {25.60} & {25.63} & {25.64} & {25.59} & {25.51} & {25.60} & {25.73} & {25.82} & {25.68} & {25.80} & \underline{25.93} & {25.90} & \textbf{26.20} \\
{Tennis Court}                 & {25.52} & {26.15} & {26.16} & {26.11} & {26.25} & {26.21} & {26.03} & {26.14} & {26.32} & {26.38} & {26.24} & {26.34} & {26.44} & \underline{26.48} & \textbf{26.71} \\
\bottomrule[1.5pt]
    \end{tabular*}
\end{table*}

\begin{table*}
    \caption{Comparison of different SR $\times$3 models on various scene classes of NWPU-RESISC45 (PSNR ($dB$)). The best and second-best results are highlighted in Bold font and underlined ones, respectively.}
    \label{table21}
    \renewcommand\arraystretch{1.1}
    \setlength\tabcolsep{3.pt}
    \centering
    \begin{tabular*}{\hsize}{@{}@{\extracolsep{\fill}}l*{15\centering}{p{5mm}}@{}}
    \toprule[1.5pt]
\multirow{2}{*}{Class} & \multirow{1}{*}{CTNet} & \multirow{1}{*}{HSENet} & \multirow{1}{*}{D-DBPN} & \multirow{1}{*}{RDN} & \multirow{1}{*}{TransE-} & \multirow{1}{*}{RCAN} & \multirow{1}{*}{SAN} & \multirow{1}{*}{MHAN} & \multirow{1}{*}{SwinIR} & \multirow{1}{*}{HAUNet} & \multirow{1}{*}{TTST} & \multirow{1}{*}{CAT} & \multirow{1}{*}{RGT} & \multirow{1}{*}{DAT} & \multirow{1}{*}{HDI-}   \\
& \multirow{1}{*}{\cite{wang2021contextual}} & \multirow{1}{*}{\cite{lei2021hybrid}} & \multirow{1}{*}{\cite{haris2018deep}} & \multirow{1}{*}{\cite{zhang2018residual}} & \multirow{1}{*}{Net\cite{lei2021transformer}} & \multirow{1}{*}{\cite{zhang2018image}} & \multirow{1}{*}{\cite{dai2019second}} & \multirow{1}{*}{\cite{zhang2020remote}} & \multirow{1}{*}{\cite{liang2021swinir}} & \multirow{1}{*}{\cite{wang2023hybrid}} & \multirow{1}{*}{\cite{xiao2024ttst}} & \multirow{1}{*}{\cite{chen2022cross}} & \multirow{1}{*}{\cite{chen2024recursive}} & \multirow{1}{*}{\cite{chen2023dual}} & \multirow{1}{*}{PRNet}   \\
\midrule[0.95pt]
{Airplane}                     & {26.46} & {26.66} & {26.98} & {26.89} & {26.95} & {26.94} & {26.92} & {27.00} & {27.06} & {27.08} & {27.08} & {27.14} & {27.12} & \underline{27.16} & \textbf{27.35} \\
{Airport}                      & {26.54} & {26.74} & {27.08} & {27.11} & {27.03} & {27.16} & {27.04} & {27.16} & {27.17} & {27.18} & {27.21} & {27.27} & {27.31} & \underline{27.34} & \textbf{27.49} \\
{Baseball Diamond}             & {27.67} & {27.89} & {28.23} & {28.22} & {28.20} & {28.29} & {28.18} & {28.28} & {28.32} & {28.33} & {28.29} & {28.38} & {28.44} & \underline{28.45} & \textbf{28.63} \\
{Basketball Court}             & {26.09} & {26.26} & {26.57} & {26.54} & {26.50} & {26.59} & {26.50} & {26.61} & {26.57} & {26.67} & {26.60} & \underline{26.77} & {26.72} & {26.69} & \textbf{26.96} \\
{Beach}                        & {27.92} & {28.01} & {28.17} & {28.25} & {28.26} & {28.33} & {28.24} & {28.31} & {28.35} & \underline{28.45} & {28.35} & {28.36} & {28.42} & {28.44} & \textbf{28.57} \\
{Bridge}                       & {29.13} & {29.46} & {29.81} & {29.80} & {29.77} & {29.81} & {29.78} & {29.83} & {29.87} & {29.99} & {29.93} & {29.98} & \underline{30.06} & {30.03} & \textbf{30.31} \\
{Chaparral}                    & {23.80} & {23.88} & {24.18} & {24.06} & {24.15} & {24.21} & {23.98} & {24.16} & {24.16} & \underline{24.27} & {24.15} & {24.20} & \underline{24.27} & {24.25} & \textbf{24.45} \\
{Church}                       & {23.78} & {23.95} & {24.25} & {24.18} & {24.22} & {24.25} & {24.17} & {24.28} & {24.29} & {24.35} & {24.25} & {24.36} & {24.41} & \underline{24.42} & \textbf{24.69} \\
{Circular Farmland}            & {29.17} & {29.36} & {29.67} & {29.68} & {29.65} & {29.76} & {29.64} & {29.77} & {29.75} & {29.83} & {29.74} & {29.83} & {29.88} & \underline{29.89} & \textbf{30.11} \\
{Cloud}                        & {29.27} & {29.30} & {29.60} & {30.37} & {30.36} & {30.41} & {30.10} & {30.20} & {30.56} & \underline{30.70} & {30.49} & {30.44} & {30.62} & {30.57} & \textbf{30.83} \\
{Commercial Area}              & {24.17} & {24.37} & {24.62} & {24.59} & {24.66} & {24.66} & {24.60} & {24.70} & {24.72} & {24.77} & {24.73} & {24.84} & {24.81} & \underline{24.86} & \textbf{25.11} \\
{Dense Residential}            & {22.94} & {23.09} & {23.35} & {23.29} & {23.39} & {23.33} & {23.31} & {23.42} & {23.41} & {23.50} & {23.41} & {23.50} & {23.47} & \underline{23.56} & \textbf{23.78} \\
{Desert}                       & {30.13} & {30.21} & {30.36} & {30.43} & {30.39} & {30.45} & {30.35} & {30.45} & {30.54} & \underline{30.60} & {30.47} & {30.58} & {30.56} & {30.59} & \textbf{30.64} \\
{Forest}                       & {26.49} & {26.56} & {26.69} & {26.69} & {26.72} & {26.72} & {26.67} & {26.75} & {26.76} & \underline{26.80} & {26.74} & {26.78} & {26.79} & \underline{26.80} & \textbf{26.87} \\
{Freeway}                      & {26.66} & {26.87} & {27.20} & {27.20} & {27.21} & {27.31} & {27.16} & {27.29} & {27.30} & {27.40} & {27.27} & {27.39} & \underline{27.46} & {27.41} & \textbf{27.60} \\
{Golf Course}                  & {29.13} & {29.35} & {29.60} & {29.68} & {29.71} & {29.74} & {29.60} & {29.77} & {29.80} & {29.87} & {29.78} & {29.84} & {29.87} & \underline{29.91} & \textbf{30.06} \\
{Ground Track Field}           & {26.23} & {26.46} & {26.78} & {26.79} & {26.78} & {26.88} & {26.77} & {26.89} & {26.87} & {26.98} & {26.95} & {26.98} & {27.04} & \underline{27.05} & \textbf{27.22} \\
{Harbor}                       & {21.73} & {21.91} & {22.35} & {22.21} & {22.34} & {22.20} & {22.18} & {22.31} & {22.36} & {22.42} & {22.39} & {22.46} & {22.48} & \underline{22.50} & \textbf{22.89} \\
{Industrial Area}              & {25.41} & {25.65} & {26.02} & {25.92} & {26.03} & {25.99} & {25.97} & {26.06} & {26.01} & {26.15} & {26.08} & {26.13} & \underline{26.20} & {26.19} & \textbf{26.53} \\
{Intersection}                 & {23.30} & {23.48} & {23.82} & {23.77} & {23.79} & {23.85} & {23.75} & {23.89} & {23.86} & {23.90} & {23.92} & {23.94} & {23.99} & \underline{24.00} & \textbf{24.27} \\
{Island}                       & {31.12} & {31.22} & {31.44} & {31.48} & {31.43} & {31.56} & {31.53} & {31.64} & {31.66} & {31.69} & {31.55} & {31.69} & \underline{31.75} & {31.74} & \textbf{31.85} \\
{Lake}                         & {29.21} & {29.29} & {29.55} & {29.57} & {29.55} & {29.61} & {29.54} & {29.60} & {29.64} & {29.64} & {29.61} & \underline{29.68} & \underline{29.68} & \underline{29.68} & \textbf{29.78} \\
{Meadow}                       & {31.08} & {31.09} & {31.25} & {31.28} & {31.35} & {31.32} & {31.25} & {31.30} & {31.32} & {31.39} & {31.31} & {31.41} & \underline{31.42} & {31.41} & \textbf{31.51} \\
{Medium Residential}           & {24.48} & {24.66} & {24.91} & {24.85} & {24.89} & {24.91} & {24.85} & {24.96} & {24.96} & {25.00} & {24.97} & {25.02} & {25.03} & \underline{25.07} & \textbf{25.25} \\
{Mobile Home Park}             & {23.10} & {23.29} & {23.64} & {23.55} & {23.63} & {23.56} & {23.58} & {23.66} & {23.64} & {23.68} & {23.72} & \underline{23.82} & {23.73} & {23.79} & \textbf{24.12} \\
{Mountain}                     & {26.74} & {26.83} & {26.92} & {26.97} & {26.97} & {27.01} & {26.93} & {27.01} & {27.02} & {27.06} & {27.03} & {27.06} & {27.07} & \underline{27.10} & \textbf{27.15} \\
{Overpass}                     & {25.16} & {25.42} & {25.87} & {25.85} & {25.84} & {25.93} & {25.79} & {25.93} & {25.91} & {26.01} & {25.98} & {26.01} & \underline{26.09} & \underline{26.09} & \textbf{26.36} \\
{Palace}                       & {23.72} & {23.91} & {24.21} & {24.15} & {24.19} & {24.26} & {24.16} & {24.29} & {24.28} & {24.32} & {24.27} & {24.35} & {24.38} & \underline{24.40} & \textbf{24.58} \\
{Parking Lot}                  & {21.62} & {21.75} & {22.07} & {21.97} & {21.99} & {21.98} & {21.96} & {22.03} & {22.05} & {22.16} & {22.14} & \underline{22.22} & \underline{22.22} & {22.21} & \textbf{22.63} \\
{Railway}                      & {25.38} & {25.59} & {25.91} & {25.93} & {25.88} & {25.97} & {25.83} & {25.98} & {25.93} & {25.99} & {25.95} & {26.03} & \underline{26.09} & \underline{26.09} & \textbf{26.32} \\
{Railway Station}              & {24.56} & {24.76} & {25.00} & {25.01} & {25.06} & {25.11} & {25.01} & {25.10} & {25.12} & {25.21} & {25.15} & {25.24} & \underline{25.26} & \underline{25.26} & \textbf{25.42} \\
{Rectangular Farmland}         & {29.54} & {29.81} & {30.08} & {30.13} & {30.11} & {30.18} & {30.08} & {30.23} & {30.25} & {30.34} & {30.24} & {30.37} & {30.41} & \underline{30.45} & \textbf{30.68} \\
{River}                        & {28.19} & {28.35} & {28.48} & {28.55} & {28.54} & {28.61} & {28.53} & {28.61} & {28.62} & {28.67} & {28.64} & {28.69} & {28.70} & \underline{28.72} & \textbf{28.83} \\
{Roundabout}                   & {24.86} & {25.03} & {25.31} & {25.29} & {25.30} & {25.37} & {25.27} & {25.39} & {25.37} & {25.44} & {25.40} & {25.43} & \underline{25.47} & \underline{25.47} & \textbf{25.66} \\
{Runway}                       & {29.18} & {29.47} & {29.88} & {29.96} & {29.93} & {30.15} & {29.91} & {30.18} & {30.13} & {30.32} & {30.19} & {30.33} & \underline{30.47} & {30.41} & \textbf{30.76} \\
{Sea Ice}                      & {26.61} & {27.01} & {27.39} & {27.25} & {27.39} & {27.29} & {27.06} & {27.27} & {27.25} & {27.55} & {27.41} & {27.50} & {27.48} & \underline{27.61} & \textbf{28.00} \\
{Ship}                         & {26.59} & {26.77} & {27.18} & {27.10} & {27.12} & {27.16} & {27.11} & {27.19} & {27.19} & {27.21} & {27.23} & {27.27} & \underline{27.31} & {27.30} & \textbf{27.46} \\
{Snowberg}                     & {22.71} & {22.78} & {23.14} & {23.08} & {23.12} & {23.14} & {23.04} & {23.18} & {23.18} & {23.22} & {23.22} & {23.24} & {23.24} & \underline{23.29} & \textbf{23.51} \\
{Sparse Residential}           & {26.41} & {26.56} & {26.85} & {26.82} & {26.83} & {26.86} & {26.80} & {26.89} & {26.89} & {26.90} & {26.91} & {26.93} & \underline{26.97} & {26.96} & \textbf{27.11} \\
{Stadium}                      & {25.26} & {25.49} & {25.78} & {25.73} & {25.75} & {25.81} & {25.75} & {25.83} & {25.86} & {25.93} & {25.94} & {25.94} & {25.98} & \underline{26.00} & \textbf{26.36} \\
{Storage Tank}                 & {24.19} & {24.36} & {24.82} & {24.72} & {24.77} & {24.78} & {24.71} & {24.83} & {24.77} & {24.93} & {24.85} & {24.94} & {24.98} & \underline{25.00} & \textbf{25.33} \\
{Tennis Court}                 & {25.67} & {25.86} & {26.11} & {26.10} & {26.10} & {26.15} & {26.06} & {26.19} & {26.17} & {26.24} & {26.20} & \underline{26.28} & {26.27} & \underline{26.28} & \textbf{26.53} \\
{Terrace}                      & {28.53} & {28.70} & {29.02} & {29.06} & {29.07} & {29.11} & {29.00} & {29.12} & {29.12} & {29.20} & {29.14} & {29.22} & {29.22} & \underline{29.27} & \textbf{29.32} \\
{Thermal Power Station}        & {26.21} & {26.39} & {26.74} & {26.75} & {26.82} & {26.86} & {26.74} & {26.86} & {26.85} & {26.91} & {26.89} & {26.93} & \underline{27.00} & {26.99} & \textbf{27.23} \\
{Wetland}                      & {28.45} & {28.56} & {28.77} & {28.81} & {28.85} & {28.85} & {28.76} & {28.82} & {28.88} & \underline{28.92} & {28.82} & {28.90} & {28.89} & {28.90} & \textbf{28.98} \\
\bottomrule[1.5pt]
    \end{tabular*}
\end{table*}

\begin{figure*}[htbp]
\centering
\includegraphics[width=7in]{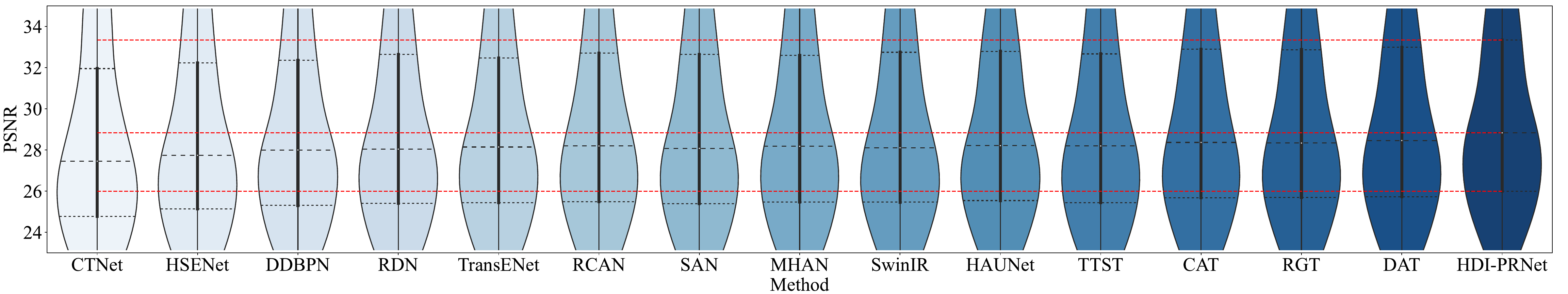}
\caption{PSNR violin-plot of different $\times$2 SR methods on WHU-RS19 dataset.}
\label{fig4}
\end{figure*}

\subsection{Comparisons with other methods on simulated datasets}
In this section, we evaluate the superiority of our proposed method by comparing it with state-of-the-art image restoration methods. Table \ref{table17} shows the quantitative analysis of the indicators of different algorithms in the $\times 2, \times 3, \times 4$ image restoration task in the WHU-RS19, DOTA, UCMerced, RSSCN7 and NWPU-RESISC45 datasets. Table \ref{table17} demonstrates that HDI-PRNet consistently achieves superior performance across all dataset tasks, achieving the highest average values of PSNR and SSIM. In particular, the proposed method shows significant improvements in reconstruction quality compared to state-of-the-art approaches, including DAT and TTST.

Tables \ref{table18}, \ref{table19}, \ref{table20}, and \ref{table21} present the PSNR results on the WHU-RS9, RSSCN7, UCMerced, and NWPU-RESISC45 datasets, demonstrating the comparative reconstruction performance in remote sensing image restoration tasks under varying sensors and land cover classes. The results reveal that HDI-PRNet achieves significantly superior reconstruction quality compared to TTST (a specialized method for remote sensing image reconstruction) and DAT (a general image super-resolution algorithm). In particular, HDI-PRNet exhibits consistent advantages across most land classes, further validating its geesubmit-neralizability and robustness in various remote sensing scenarios. These results highlight the method's ability to deliver high-fidelity reconstructions while adapting to varying imaging conditions and scene complexities.

Fig. \ref{fig3} presents a qualitative comparison of the restoration results in different methods. Our approach significantly mitigates image degradation caused by noise, blur, and downsampling, producing sharper and more visually realistic reconstructions. Although existing methods demonstrate certain improvements, they still struggle to fully recover fine textures and structural details. For instance, CTNet, HSENet, and TransENet exhibit noticeable artifacts and over-smoothing in their outputs. Although more sophisticated architectures such as DAT and HAUNet leverage richer feature representations, they fail to reconstruct edge details with sufficient clarity. In contrast, our method not only suppresses noise effectively, but also preserves and enhances critical image details and edges, yielding the clearest and most accurate remote sensing reconstructions among all compared approaches.

Fig. \ref{fig4} shows a violin plot that serves as a tool to evaluate the performance stability of the HDI-PRNet restoration framework. The plot includes three primary lines: the first quartile (Q1), the median, and the third quartile (Q3), which are useful for assessing the result's dispersion. The Q1, median and Q3 values of the proposed method are the highest, indicating better overall performance compared to other approaches. As depicted in Fig. \ref{fig4}, our method exhibits better reconstruction performance and can handle different data more stably.

\begin{figure*}[htbp]
\centering
\includegraphics[width=6in]{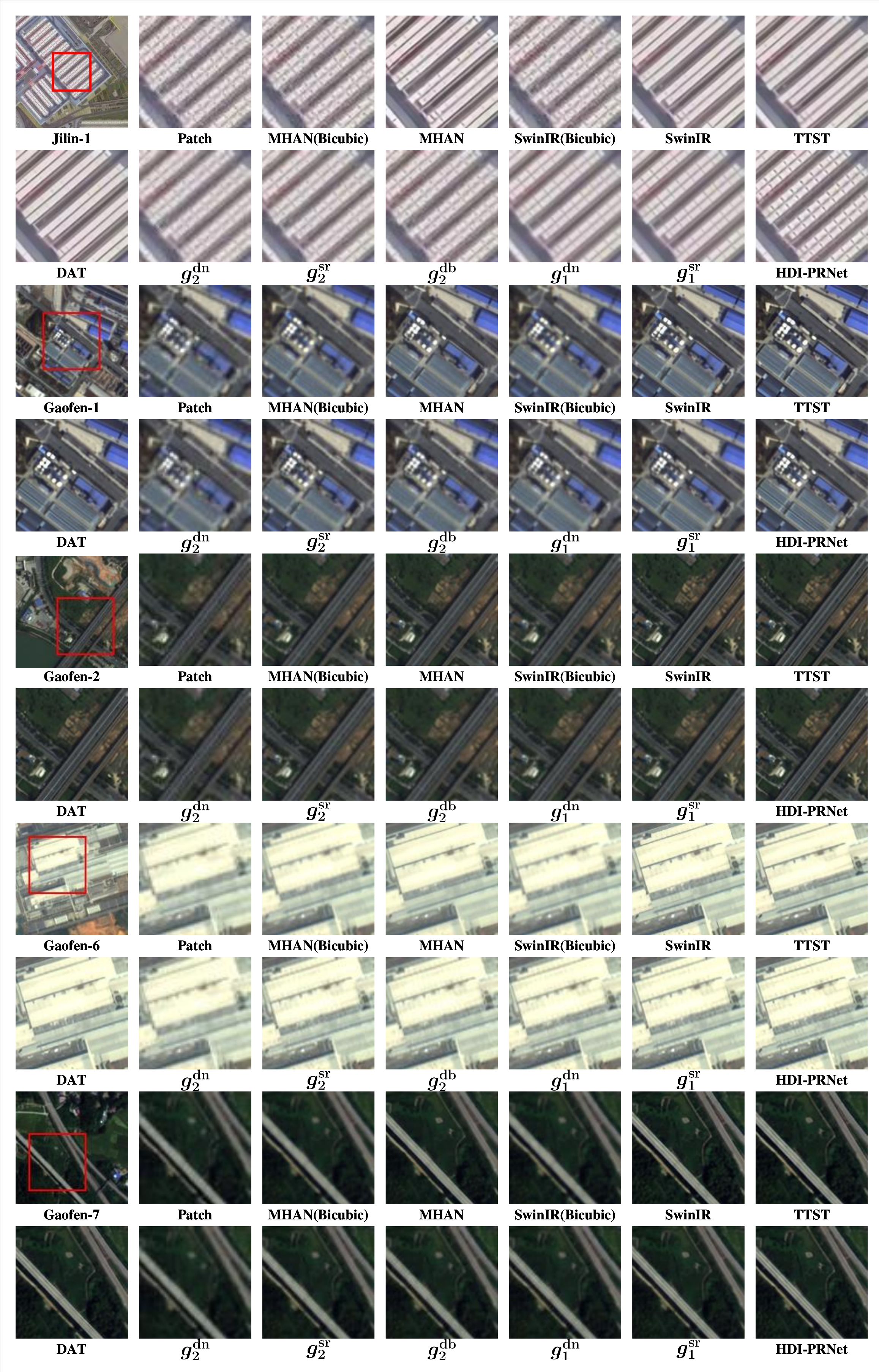}
\caption{Visualization comparison of different methods on the real satellite images.}
\label{fig5}
\end{figure*}

\begin{table*}
    \caption{Comparison of different restoration models on real satellite images.}
    \label{table10}
    \renewcommand\arraystretch{1.1}
    \setlength\tabcolsep{2pt}
    \centering
    \begin{tabular*}{\hsize}{@{}@{\extracolsep{\fill}}ccccccccccc@{}}
    \toprule[1.5pt]
BRISQUE    & HSENet & \multicolumn{1}{c}{TransENet} & \multicolumn{1}{c}{MHAN} & \multicolumn{1}{c}{MHAN (Bicubic)} & \multicolumn{1}{c}{SwinIR} & \multicolumn{1}{c}{SwinIR (Bicubic)} & \multicolumn{1}{c}{TTST} & \multicolumn{1}{c}{HAUNet} & \multicolumn{1}{c}{DAT}  & \multicolumn{1}{c}{HDI-PRNet}\\
\midrule[0.95pt]
Jilin-1  & 41.3774 & 41.1180 & \textbf{35.8704} & 43.1953 & 37.1765 & 43.6537 & 37.3207 & 43.4201 & 36.8362 & \underline{36.5810}  \\
\cmidrule(r){1-11}
Gaofen-1 & 40.3188 & \underline{39.6849} & 40.2469 & 45.5903 & 41.4091 & 45.4889 & 40.9722 & 41.3830 & 40.7827 & \textbf{39.6351}  \\
\cmidrule(r){1-11}
Gaofen-2 & \underline{37.3217} & 38.3410 & 37.9373 & 39.8892 & 39.3446 & 39.4106 & 39.2308 & 39.1100 & 38.5144 & \textbf{35.9592}  \\
\cmidrule(r){1-11}
Gaofen-6 & \underline{36.8403} & 38.3920 & 38.3865 & 41.4711 & 38.9494 & 41.1936 & 38.7809 & 38.9542 & 38.4101 & \textbf{36.7054} \\
\cmidrule(r){1-11}
Gaofen-7 & 37.7147 & 38.0881 & 37.7705 & 43.6968 & 38.6543 & 43.0609 & 38.3997 & 37.8802 & \underline{37.1173} & \textbf{37.0412} \\
\bottomrule[1.5pt]
    \end{tabular*}
\end{table*}

\subsection{Comparisons on real remote sensing images}
To validate the generalizability of the proposed method, we perform experiments using real remote sensing images captured by the Jilin-1, Gaofen-1, Gaofen-2, Gaofen-6, and Gaofen-7 satellites. To facilitate testing, the dataset includes $824$ Jilin-1 images that have been resized to $64 \times 64$ pixels, along with $400$ images from each Gaofen satellite resized to $128 \times 128$ pixels.

The visualization comparison is shown in Fig.\ref{fig5}. MHAN, SwinIR, DAT, and TTST have evidently enhanced image quality to a certain extent; however, they have excessively smoothed the original image, unintentionally removing some details by mistaking them for noise or artifacts. In comparison to other methods, the reconstructed image using the proposed method is significantly clearer, particularly in detail and texture, restoring these aspects more accurately and closely to the original image without producing artifacts or over-smoothing. Furthermore, we train the MHAN and SwinIR models employing bicubic degradation and evaluate them on real remote sensing datasets. Both MHAN (bicubic) and SwinIR (bicubic) demonstrate a slight but not evident improvement in the quality of real remote sensing images,  which further demonstrates the rationality of our model's architecture and configurations.

Since real remote sensing images lack HR ground truth, we employ BRISQUE to evaluate the predicted results. As shown in Table \ref{table10}, HDI-PRNet achieves the best or second-best results on real images from five different satellites. Although MHAN records the highest score on Jilin-1, its predictions are visually over-smoothed, which is not preferable. Overall, these significant improvements in both visual quality and reconstruction performance metrics strongly show the effectiveness and practical utility of HDI-PRNet to restore real remote sensing images, further highlighting the excellence of our proposed HDI-PRNet.

\section{Conclusion}\label{sec05}
In this paper, we proposed a novel remote sensing high-order restoration framework. Our framework is based on the inverse Markov process design of the high-order degradation model to gradually restore the image, which provides a theoretical basis for the design of the DL framework. In addition, we introduced a denoising module that simulates the proximal mapping operator to learn the denoising prior representation of the current image. To solve the deblurring problem, we proposed a deblurring module that ensures the accurate representation of the degradation operator based on the expansion of the Neumann series and dual-domain degradation learning. We also introduced an intermediate loss in the middle of the network so that each module has supervision and the overall network is more interpretable. We performed several high-order degradation reconstruction experiments on WHU-RS19 and real remote sensing images. Experimental results show that our HDI-PRNet achieves competitive results on both synthetic and real images.

Although HDI-PRNet has achieved excellent performance in the task of real remote sensing image restoration, there are still some limitations. To verify the effectiveness of the high-order model design, this paper only discussed the three classic image restoration tasks of denoising, deblurring, and super-resolution, but the real remote sensing imaging process also involves various other types of degradation, such as clouds and artifacts. The high-order degradation model composed of more types of degradation makes the restoration process of real images complex and challenging.

In the future, we believe that our method can be extended to other image restoration problems to perform more complex and diverse image restoration tasks, such as image dehazing and image artifact removal, and obtain better restoration performance on real images.
\bibliographystyle{IEEEtran}
\bibliography{DL_SR_ref-resubmission}

\end{document}